\documentclass[final]{ieeeaccess}
\usepackage{cite}
\usepackage{amsmath,amssymb,amsfonts}
\usepackage{algorithm}
\usepackage{booktabs}

\usepackage{graphicx}
\usepackage{bm}
\usepackage{booktabs}
\usepackage{subcaption}
\usepackage{float}
\usepackage{algpseudocode}
\usepackage{url}

\newcommand{\citet}[1]{\cite{#1}}
\newcommand{\citep}[1]{\cite{#1}}
\newtheorem{theorem*}{Theorem}
\newtheorem{proof}{Proof}

\makeatletter
\def\headerlogo{}%
\def\headerlogoall{}%
\makeatother
\makeatletter
\def\ps@headings{%
  \def\@oddhead{}%
  \def\@evenhead{}%
  \def\@oddfoot{\hfil\thepage\hfil}
  \def\@evenfoot{\hfil\thepage\hfil}
}
\def\ps@titlepage{%
  \def\@oddhead{}%
  \def\@evenhead{}%
  \def\@oddfoot{\hfil\thepage\hfil}%
  \def\@evenfoot{\hfil\thepage\hfil}%
}
\makeatother
\begin{document}

\doi{10.1109/ACCESS.2026.3686867}
\title{NABLA: Neighborhood Adaptive Block-Level Attention for Efficient Video Generation}
\author{
Dmitrii Mikhailov\authorrefmark{1,*},
Aleksey Letunovskiy\authorrefmark{1},
Maria Kovaleva\authorrefmark{1},
Vladimir Arkhipkin\authorrefmark{1},\\
Vladimir Korviakov\authorrefmark{1,*},
Vladimir Polovnikov\authorrefmark{1,2,*},
Viacheslav Vasilev\authorrefmark{1,3},
Evelina Sidorova\authorrefmark{1},\\
Denis Dimitrov\authorrefmark{1}
}
\address[1]{Kandinsky Lab, Moscow, Russia}
\address[2]{Lomonosov Moscow State University (MSU), Moscow, Russia}
\address[3]{Moscow Center for Advanced Studies, Moscow, Russia}
\corresp{*Corresponding authors: \{dmitriy.mikhaylov, vladimir.korviakov, vladimir.polovnikov\}@kandinskylab.ai}

\begin{abstract}
Full self-attention in video diffusion transformers scales quadratically with the spatio-temporal token count, making processing high-resolution video generation prohibitively slow and memory-intensive. We introduce \textbf{NABLA}, a Neighborhood-Adaptive Block-Level Attention mechanism that constructs a per-head sparse mask in three steps: (i) average-pooling queries and keys into $N \times N$ blocks, (ii) retaining the highest-probability blocks via a cumulative distribution function (CDF) threshold, and (iii) optionally unioning the result with Sliding-Tile Attention (STA) to mitigate boundary artifacts. NABLA integrates seamlessly into PyTorch's FlexAttention without requiring custom kernels or auxiliary losses. Extensive experiments demonstrate significant acceleration for both training and inference: on the Wan 2.1 14B text-to-video model at 720p, NABLA achieves up to $2.7\times$ speed-up in inference while matching baseline quality metrics (CLIP: $42.06 \rightarrow 42.08$, VBench: $83.16 \rightarrow 83.17$, FVD: $68.9 \rightarrow 67.5$). Furthermore, during pre-training of a 2B DiT at $512^2$ resolution, NABLA reduces iteration time from 10.9s to 7.5s ($1.46\times$ acceleration) while achieving lower validation loss.
\end{abstract}

\begin{keywords}
diffusion models, efficient attention, sparse attention, transformer acceleration, video generation
\end{keywords}

\titlepgskip=-21pt
\maketitle
\begingroup
\footnotesize
\noindent \copyright~2026 IEEE. Personal use of this material is permitted. Permission from IEEE must be obtained for all other uses, in any current or future media, including reprinting/republishing this material for advertising or promotional purposes, creating new collective works, for resale or redistribution to servers or lists, or reuse of any copyrighted component of this work in other works.

\noindent This is the author's version of an article that has been published in IEEE Access. The final version of record is available at:\\ \texttt{https://doi.org/10.1109/ACCESS.2026.3686867}
\endgroup

\begin{figure}[!htb]
\centering
\includegraphics[width=1.0\linewidth]{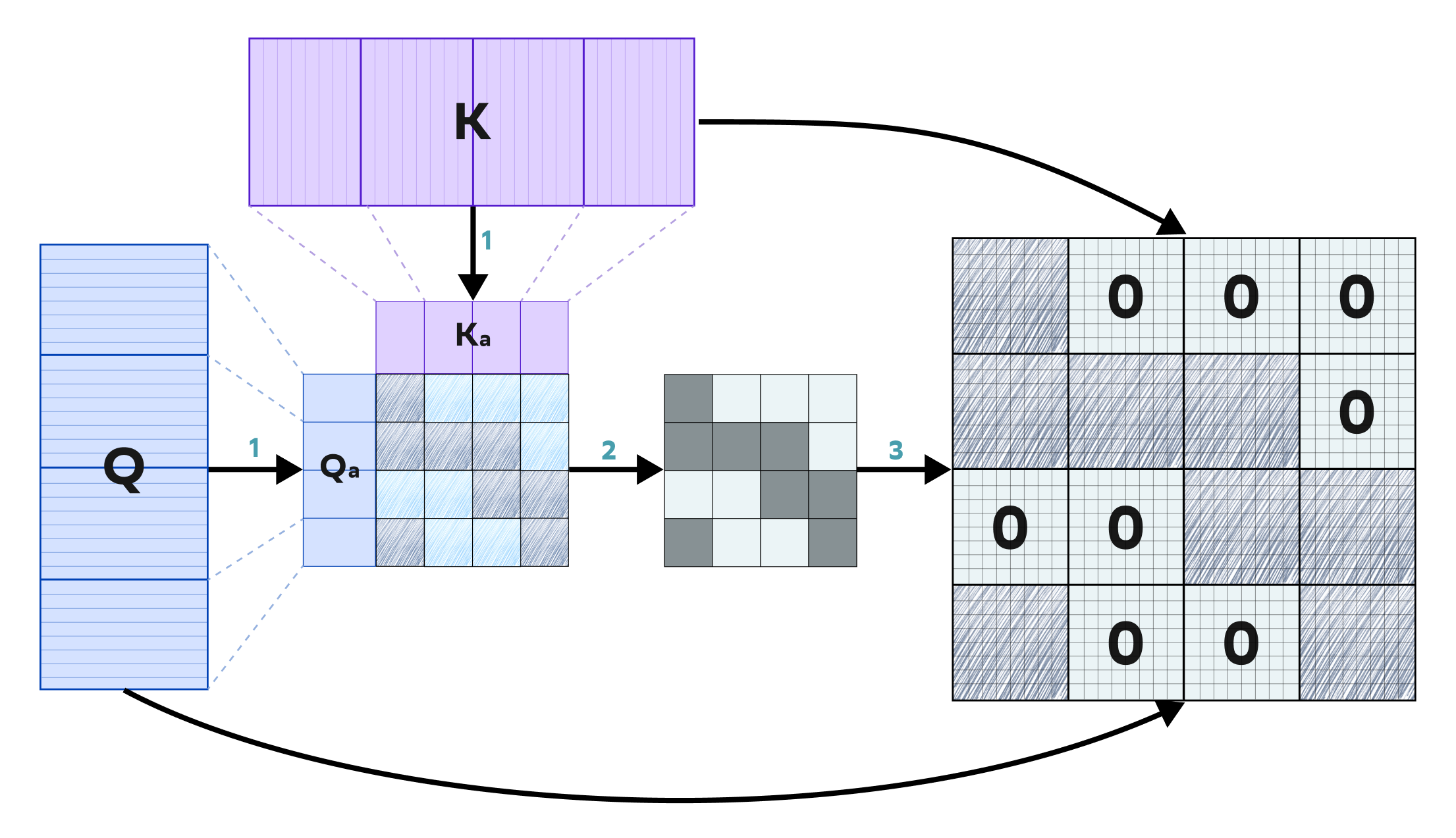}
\caption{The block-sparse attention mask is computed by (1) reducing the dimensionality of queries (Q) and keys (K), (2) sparsifying the softmax distribution via a cumulative density function (CDF) threshold and binarizing the result, and (3) mapping the sparse mask back to the original input blocks.}
\label{fig:nabla}
\end{figure}
\section{Introduction}
\label{sec:introduction}

Among various generative methods, diffusion~\cite{diffusion2020ho} is currently the state-of-the-art approach for generating media content such as images and videos. One of the major milestones in the development of this approach was the introduction of the diffusion transformer. It was first proposed for image generation in~\cite{Peebles2022DiT}, where the authors demonstrated that this architecture surpasses the previously dominant U-Net frameworks for this task and also highlighted its scalability.
Another key advancement in modern content generation using diffusion methods is latent diffusion~\cite{rombach2021highresolution}, where the diffusion process operates not on raw images or videos but on their compressed representations obtained through variational autoencoders~\cite{Kingma2014VAE}.
This is particularly important for video generation due to computational complexity.
Diffusion transformers can be broadly categorized into two main classes: CrossDiT and MMDiT~\cite{flow2024esser}. The key differences between these architectures lie in their handling of text embeddings and the attention~\cite{vaswani2017attention} mechanisms employed. In CrossDiT, text tokens are processed separately from visual tokens and incorporated via cross-attention. In contrast, MMDiT processes text and visual tokens in parallel and blends them through self-attention. This architectural distinction may influence the structure of the attention matrices learned by diffusion transformers.

The full potential of transformers in latent diffusion for video generation was first demonstrated by the closed-source solution Sora~\cite{sora2024openai}.
Since then, numerous popular closed-source and open-source solutions have emerged, such as MovieGen~\cite{polyak2025moviegencastmedia}, HunyuanVideo~\cite{kong2025hunyuanvideo}, CogVideoX~\cite{yang2025cogvideox}, Kling~\cite{kling2024}, WAN~\cite{wan2025} and Kandinsky~\cite{10815947, vladimir-etal-2024-kandinsky}.
While these models advance video generation capabilities, they share a critical limitation: computationally expensive full attention mechanisms.
However, recent theoretical analyses~\cite{deng2025sparse} and experimental evidence~\cite{tan2024dsv, xia2025trainingfree, zhang2025spargeattn, jiang2024minference} have revealed that attention matrices demonstrate inherent sparsity patterns, suggesting significant potential for optimizing computational efficiency through sparse attention mechanisms.

The methods for simplifying attention mechanisms in video generation models presented in the literature can be divided by two main features: dynamism and usage scenario. The first feature determines how important patterns are distinguished in attention masks, statically or dynamically. The second feature implies whether the proposed method can be used in zero-shot mode or the model should be trained with it.
The most important and oldest static patterns in attention masks include sliding window attention~\cite{Beltagy2020Longformer} in NLP tasks and window-based attention (SWIN)~\cite{liu2021Swin} in CV tasks. In recent years, neighbor attention (NATTEN)~\cite{hassani2023neighborhood, hassani2022dilated} as well as its effective implementations~\cite{hassani2024faster} has emerged as the successor to SWIN attention. It uses close idea but with intersecting windows. Recent work Sliding Tile Attention~\cite{zhang2025fast} builds upon NATTEN principles but optimizes it for efficient computations on GPU using the correct size of visual block.
However, although static patterns use empirical knowledge about the data structure in various domains such as NLP or CV, they do not always correspond to the patterns that real transformers obtain during training.
Moreover, many works~\cite{tan2024dsv, xia2025trainingfree, xi2025sparse, jiang2024minference, wen2025analysis} have shown that attention masks differ across blocks, heads, text prompts, and even at different steps of video generation.
To address this, several dynamic approaches have been proposed. For instance, MInference~\cite{jiang2024minference} identifies groups of attention masks offline and selects parameters during inference, while Sparse VideoGen~\cite{xi2025sparse} employs online profiling to classify heads into spatial or temporal types. Other methods like AdaSpa~\cite{xia2025trainingfree} utilize dynamic online search, and SpargeAttn~\cite{zhang2025spargeattn} approximates masks on compressed queries and keys. However, these dynamic methods often rely on predefined pattern dictionaries or require additional profiling stages, which may limit their adaptability to diverse video content dynamics and introduce latency overhead.

While the aforementioned works primarily focus on accelerating inference for pre-trained video generation models, similar concepts have also been applied to model pre-training and fine-tuning. Numerous studies have explored sparse or simplified attention in NLP tasks~\cite{NEURIPS2023_bc222e81, kitaev2020reformer, wang2020linformer, goncalves2025adasplash, willette2025delta}.
In Native Sparse Attention~\cite{Yuan2025NativeSA}, the authors reduce the number of keys and values in LLMs in various ways while maintaining the size of the query.
Three main patterns stand out: compression, selection, and sliding window, all of which are applicable to both inference and training.
An important work on training a video generator with sparse attention is DSV~\cite{tan2024dsv}.
It employs a two-stage training. At the first stage, a low-rank predictor for the attention matrix and a sparsity estimator are trained. At the second stage, the model is trained with the predictor and estimator held fixed.

In summary, numerous methods exist for accelerating attention mechanisms within transformer models. This area has been particularly well-explored in the field of Natural Language Processing. For video generation, this area is still developing. Although several approaches focus exclusively on inference or, at best, fine-tuning, video generation models pretrained with sparse attention remain limited.

In this work, we present \textbf{NABLA} (Neighborhood-Adaptive Block-Level Attention), a novel mechanism that combines the simplicity and strong priors of Sliding Tile Attention (STA)~\cite{zhang2025fast} with the flexibility of training-based approaches. NABLA utilizes a simple downsampling strategy to construct content-aware sparsity masks dynamically via cumulative distribution function (CDF) thresholding, eliminating the need for additional training stages or complex profiling. 
To contextualize NABLA within the broader landscape of efficient multimodal systems, it is important to distinguish between computational and memory optimization. While recent works like CoMem~\cite{zhou2026comem} address \textit{memory bottlenecks} and catastrophic forgetting in continual learning via concept-graph rehearsal, NABLA specifically targets the \textit{computational complexity} (FLOPs) of the attention mechanism during generation. These approaches are complementary: combining NABLA's compute-efficient attention with memory-centric frameworks represents a promising direction for deploying large-scale video models on resource-constrained devices.

Our key contributions include:
\begin{itemize}
    \item \textbf{Efficient content-aware mask construction}, outperforming fixed sparse patterns like STA (validated experimentally).
    \item \textbf{Complementarity with STA} and other acceleration techniques.
    \item \textbf{Simple implementation} via FlexAttention without custom CUDA kernels.
    \item \textbf{Acceleration of both inference and training} of DiT models due to the fast online algorithm.
\end{itemize}

An overview of the NABLA mechanism is provided in Figure~\ref{fig:nabla}. Extensive evaluations on video datasets demonstrate that NABLA maintains performance equivalent to full attention across VBench, CLIP, FVD, and human evaluation metrics. Specifically, our experiments show a $2.7\times$ speed-up in inference (on Wan 2.1 14B at 720p) and a $1.46\times$ acceleration in pre-training (on a custom 2B DiT at $512^2$), achieved without excessive additional overhead in the training and inference pipelines. 
\section{Background}
\begin{figure*}[ht]
\centering
\begin{subfigure}{0.24\linewidth}
\includegraphics[width=\linewidth]{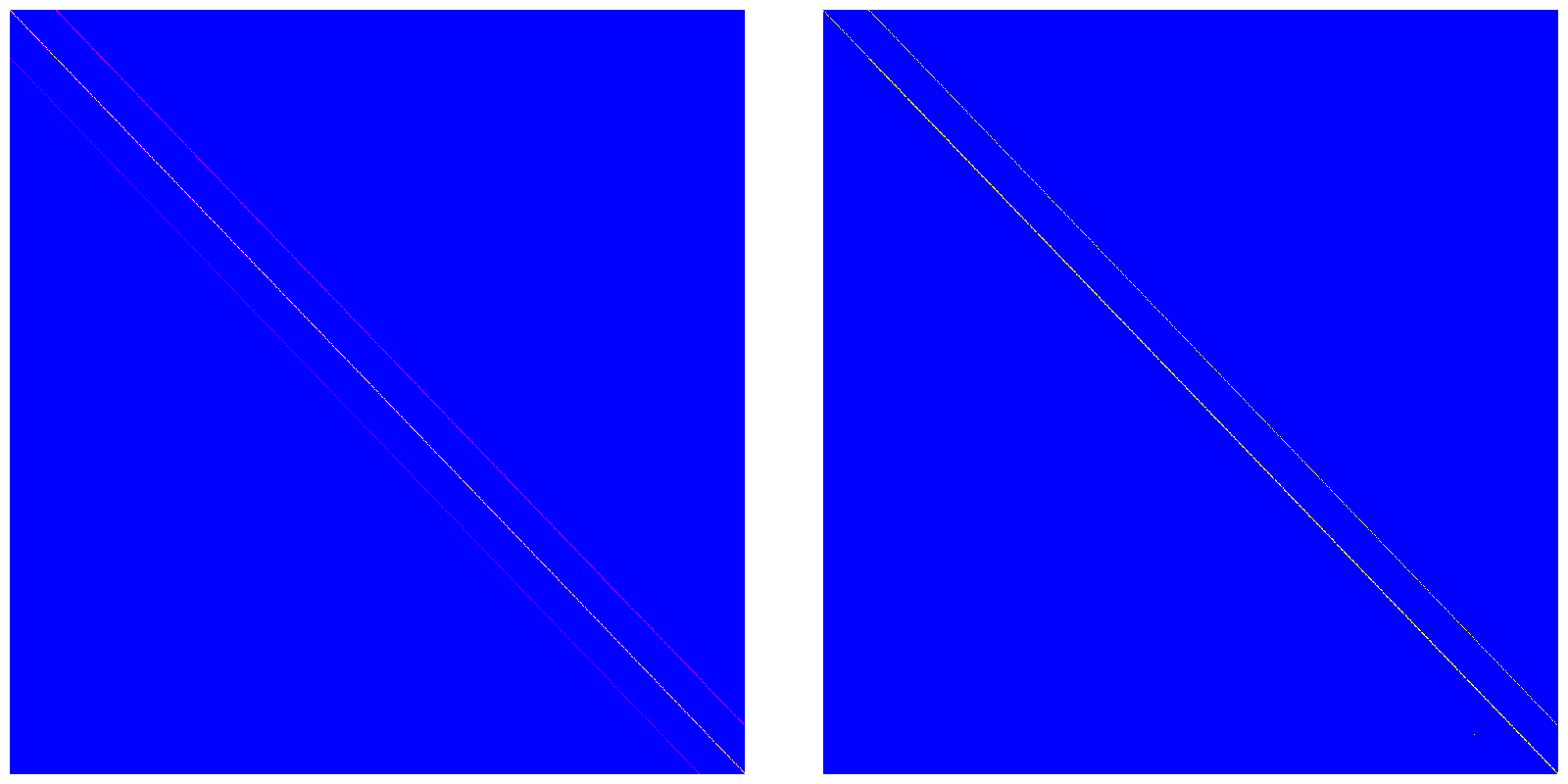}
\caption{ }
\label{fig:a}
\end{subfigure}%
\hfill
\begin{subfigure}{0.24\linewidth}
\includegraphics[width=\linewidth]{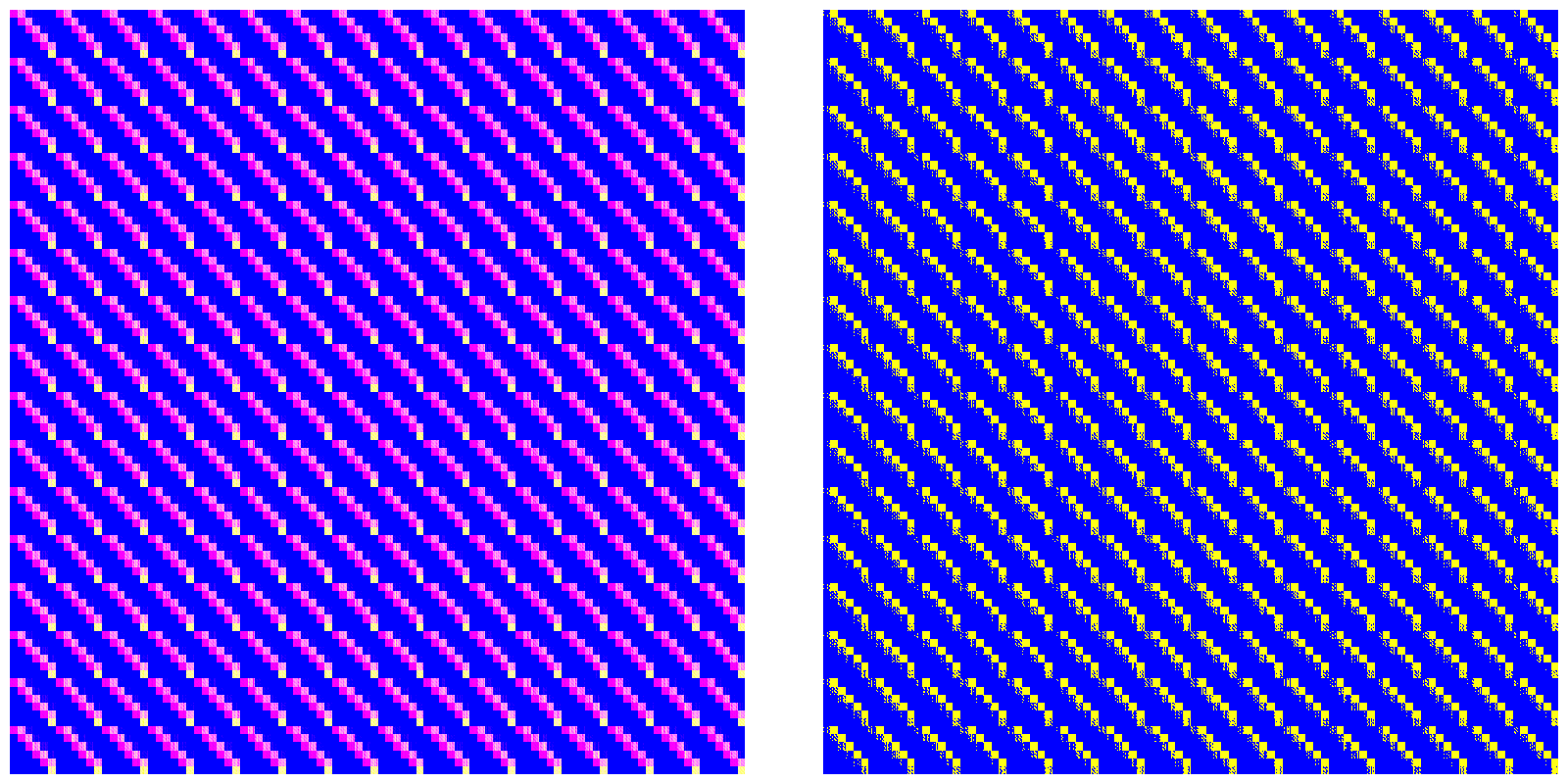}
\caption{ }
\label{fig:b}
\end{subfigure}
\hfill
\begin{subfigure}{0.24\linewidth}
\includegraphics[width=\linewidth]{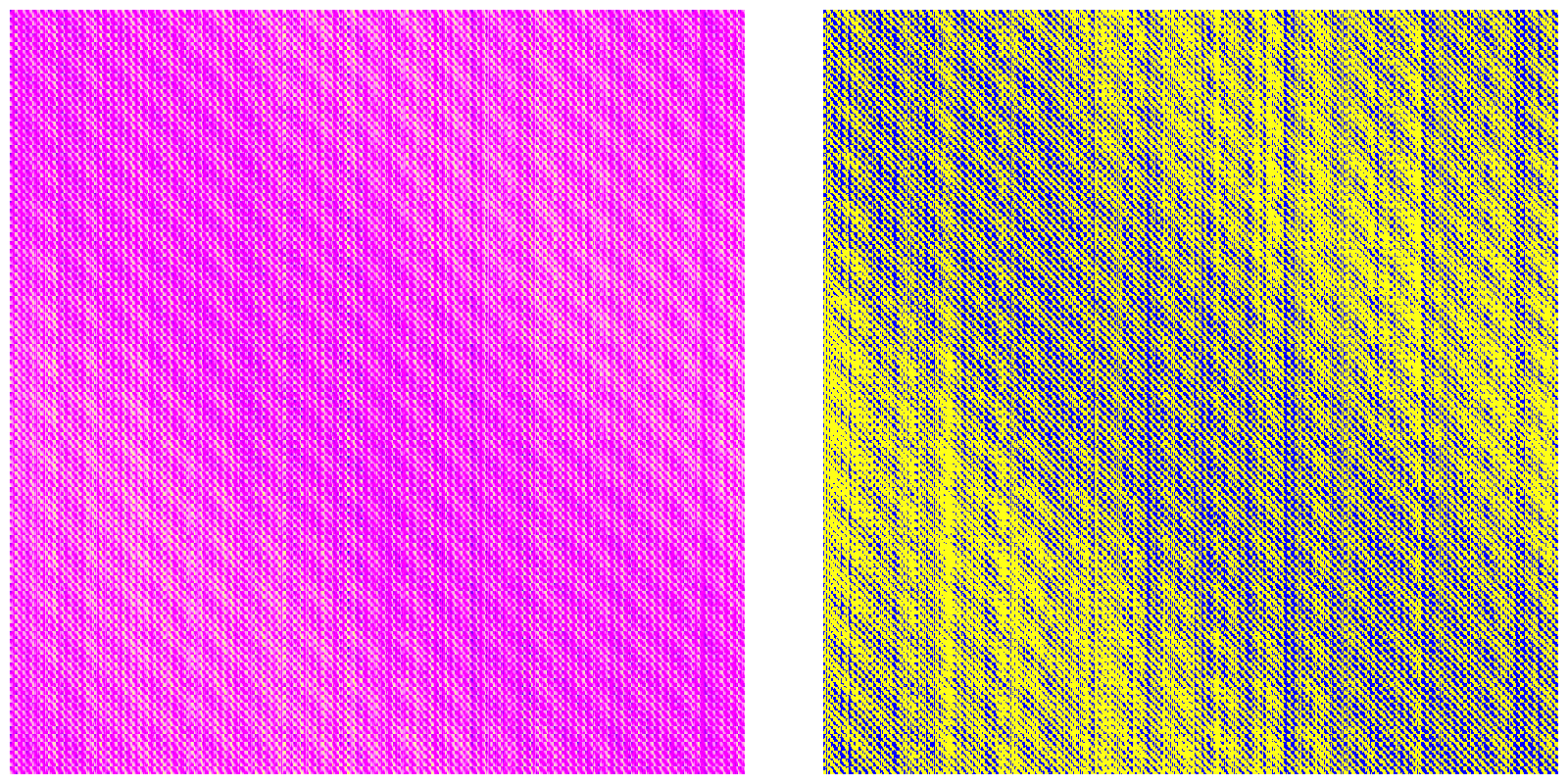}
\caption{ }
\label{fig:c}
\end{subfigure}
\hfill
\begin{subfigure}{0.24\linewidth}
\includegraphics[width=\linewidth]{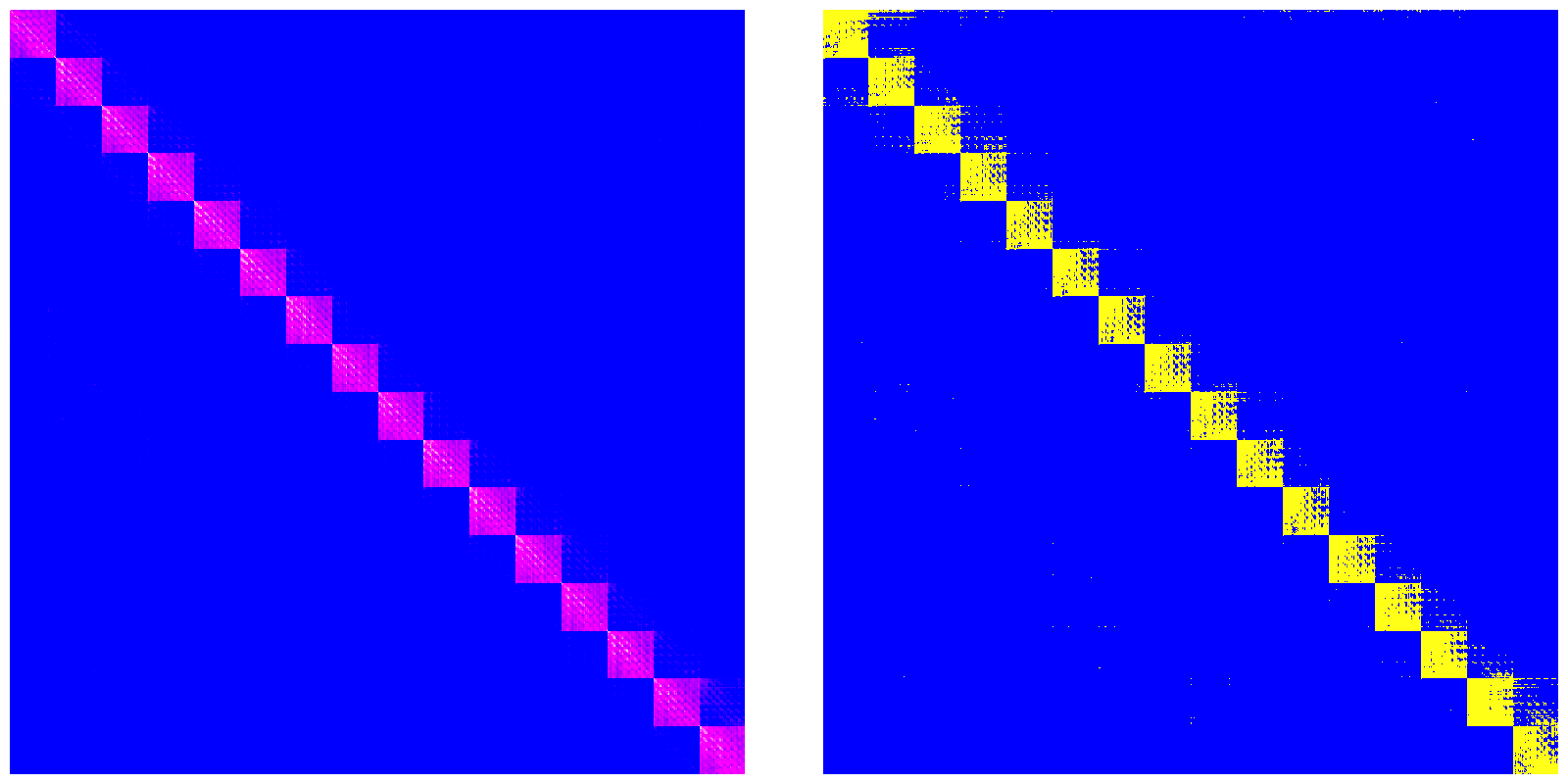}
\caption{ }
\label{fig:d}
\end{subfigure}
\hfill
\begin{subfigure}{0.24\linewidth}
\includegraphics[width=\linewidth]{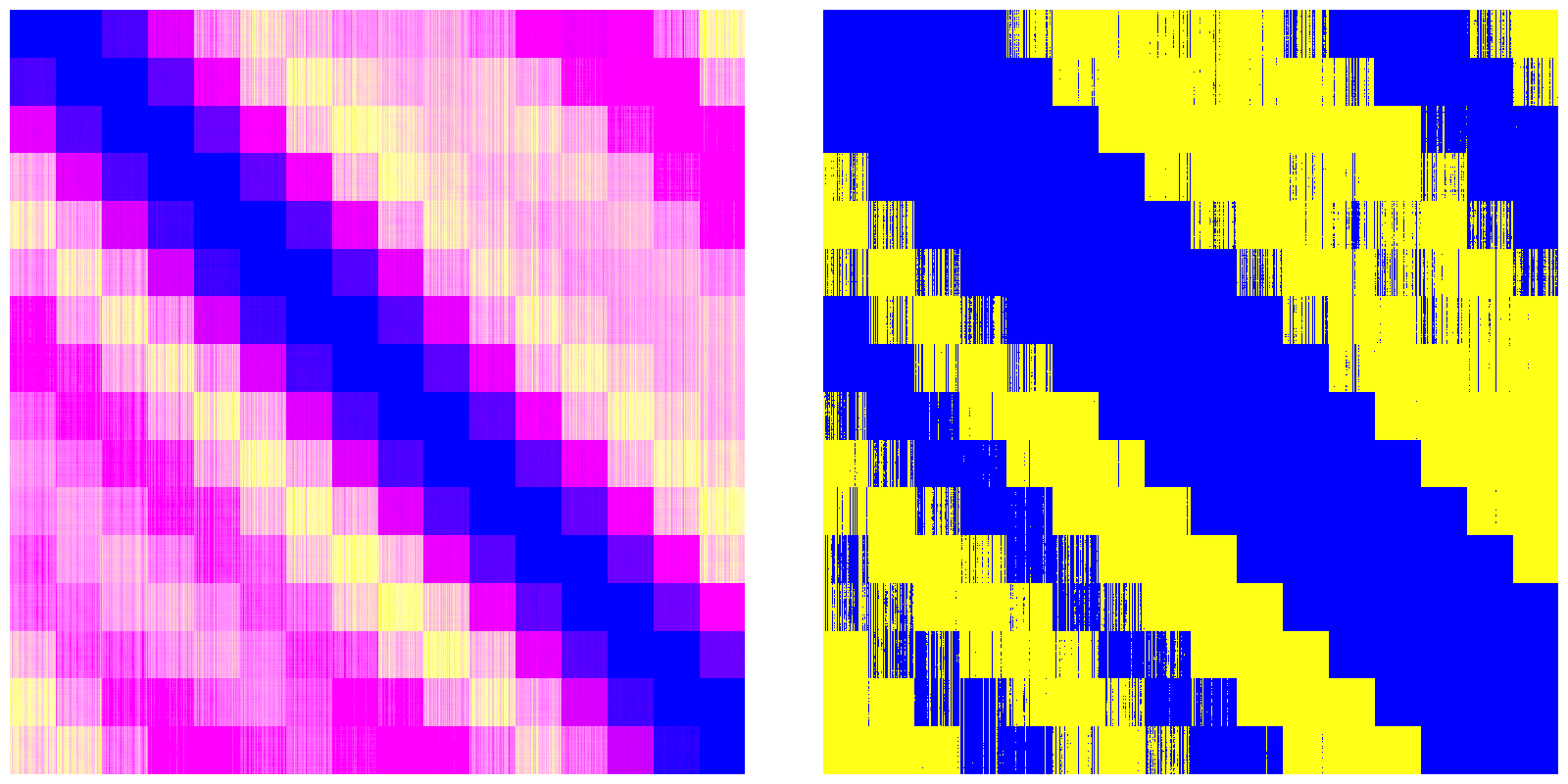}
\caption{ }
\label{fig:e}
\end{subfigure}
\hfill
\begin{subfigure}{0.24\linewidth}
\includegraphics[width=\linewidth]{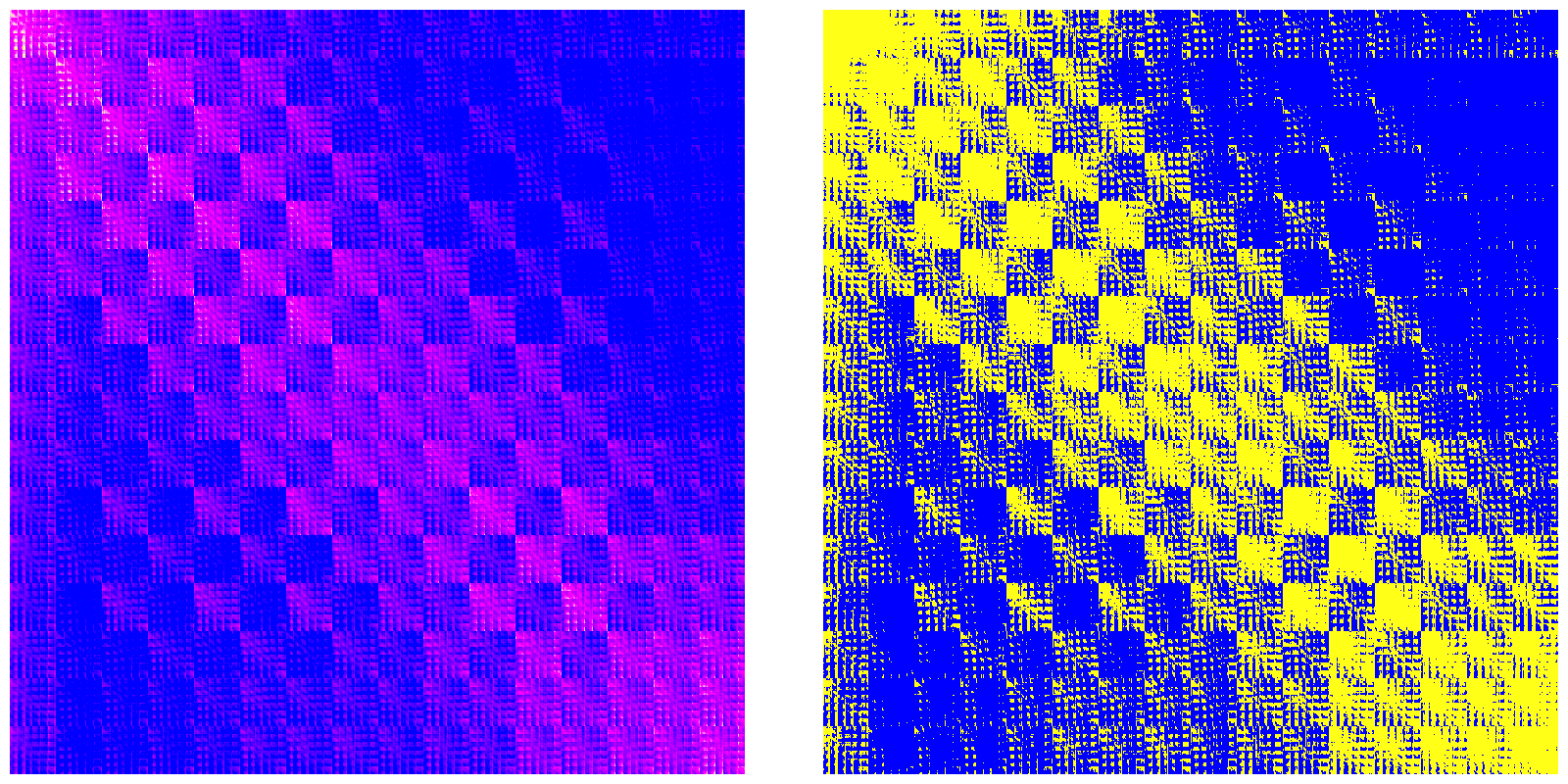}
\caption{ }
\label{fig:f}
\end{subfigure}
\hfill
\begin{subfigure}{0.24\linewidth}
\includegraphics[width=\linewidth]{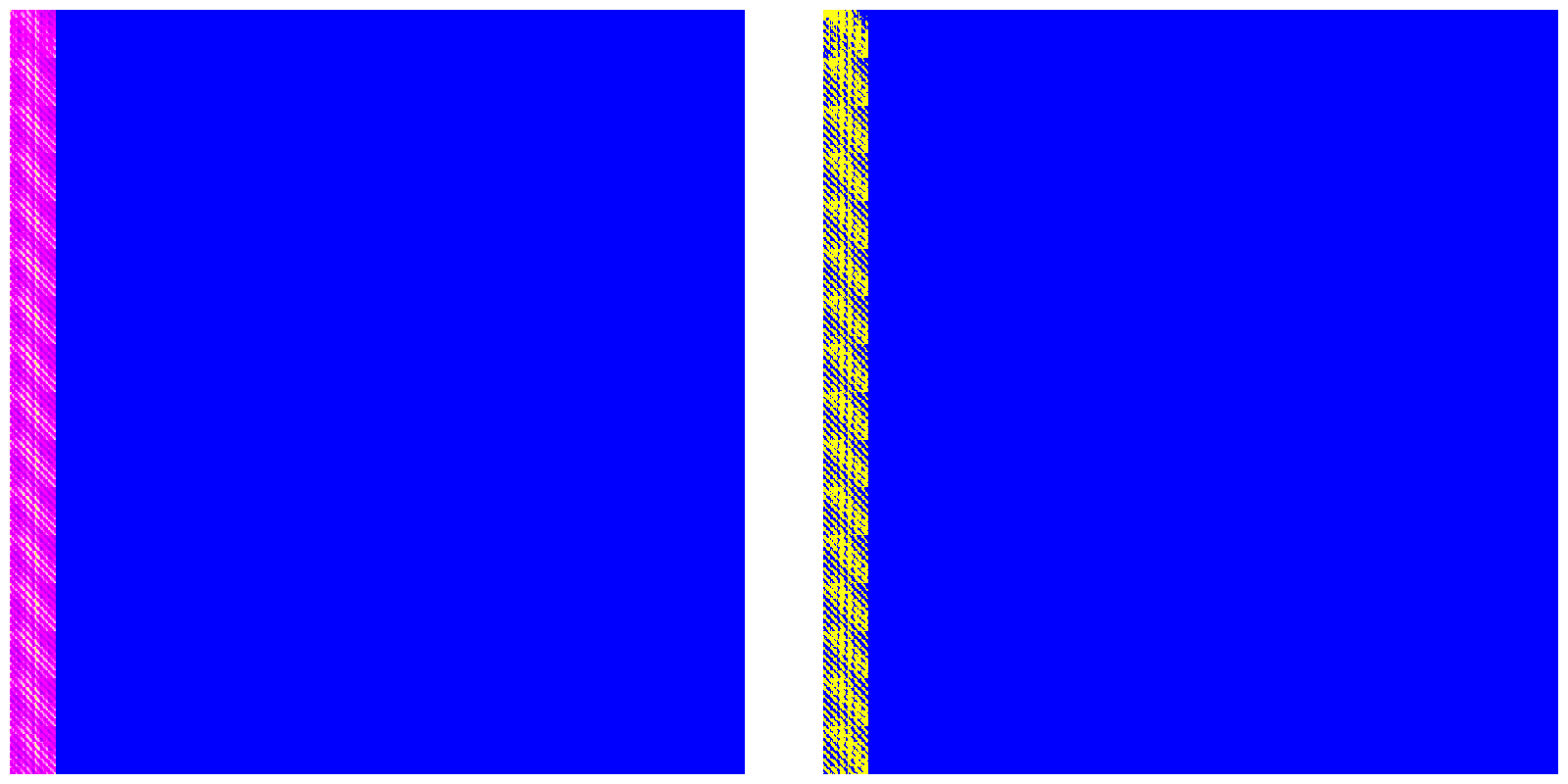}
\caption{ }
\label{fig:g}
\end{subfigure}
\hfill
\begin{subfigure}{0.24\linewidth}
\includegraphics[width=\linewidth]{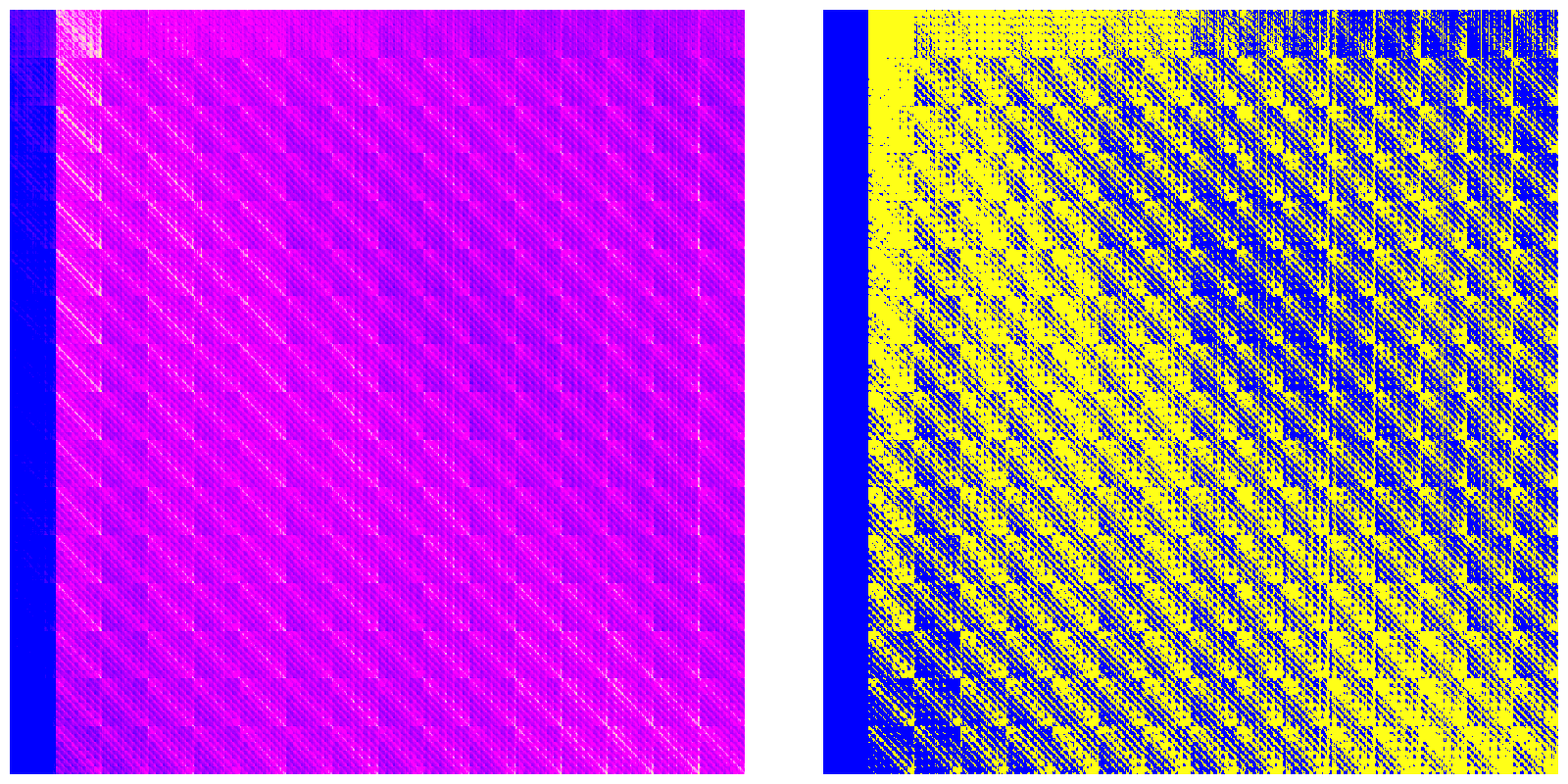}
\caption{ }
\label{fig:h}
\end{subfigure}
\caption{Examples of attention maps (left) and corresponding NABLA masks (right) for different heads of Wan 2.1 14B T2V layers.}
\label{fig:attn_maps}
\end{figure*}
\subsection{Attention Mechanism in Visual Domain}
The classical self-attention mechanism, introduced in~\cite{vaswani2017attention}, revolutionized deep learning by enabling dynamic focus on relevant parts of input data through pairwise token interactions. In visual domains, this mechanism processes images and videos by first dividing them into patches and projecting them into an embedding space \cite{dosovitskiy2021imageworth16x16words}. For an input sequence $X \in \mathbb{R}^{S \times D}$, where $S$ is the number of tokens (e.g., image patches or spatio-temporal video blocks) and $D$ is the embedding dimension, the self-attention mechanism projects these tokens into queries $Q$, keys $K$, and values $V$ using learnable weight matrices $W_Q, W_K, W_V \in \mathbb{R}^{D \times D}$:
\begin{equation}
Q = XW_Q, \quad K = XW_K, \quad V = XW_V.
\end{equation}
Attention scores are computed using a scaled dot product between queries and keys, followed by a softmax operation to produce the attention matrix \( A \in \mathbb{R}^{S \times S} \).
\begin{equation}
A = \text{softmax}\left( \frac{QK^T}{\sqrt{D}} \right),
\end{equation}
where each entry $A_{ij}$ determines how much token $j$ influences token $i$. The final output is a weighted sum of values based on these scores:
\begin{equation}
\text{Output} = AV.
\end{equation}
For video diffusion transformers (DiTs) \cite{Peebles2022DiT}, the input consists of spatio-temporal tokens (e.g., $X \in \mathbb{R}^{T \times H \times W \times D}$, where $T$ is the number of frames and $H, W$ are spatial dimensions of the latent space). In this case, the classical self-attention approach faces significant challenges due to the quadratic complexity $\mathcal{O}((T \cdot H \cdot W)^2)$ relative to the number of tokens. High-resolution or long-duration videos exacerbate this issue, as the sequence length grows cubically with spatial and temporal dimensions. In addition, many elements of attention map are near zero due to locality in space and time, which leads to redundant computations. Figure~\ref{fig:attn_maps} contains examples of attention maps in which computations on only significant elements have less complexity: $\mathcal{O}(T \cdot H \cdot W)$ (~\ref{fig:a}), $\mathcal{O}(T^2 \cdot H \cdot W)$ (~\ref{fig:b}, ~\ref{fig:h}), $\mathcal{O}(T \cdot (H \cdot W)^2)$ (~\ref{fig:d}, ~\ref{fig:f},  ~\ref{fig:g}). Although full attention theoretically preserves global coherence, its computational cost becomes prohibitive, creating a critical bottleneck for practical applications.
\subsection{Sliding Tile Attention (STA)}
To address the limitations of full attention in video generation models, Sliding Tile Attention (STA) \cite{zhang2025fast} was proposed as a hardware-efficient alternative that leverages the inherent 3D locality observed in pretrained video DiTs. STA organizes an input video latent $X \in \mathbb{R}^{T \times H \times W}$ (flattened into $S = THW$ tokens) through a tiling mechanism. Tokens are partitioned into non-overlapping tiles of size $(B_T, B_H, B_W)$, which matches GPU block sizes (for FlashAttention compatibility). STA computes attention only between query tiles and key tiles within a fixed 3D window of size $(W_T, W_H, W_W)$ centered on the query tile:
\begin{equation}
A = \text{softmax}(\frac{QK^\top}{\sqrt{D}} + M), \quad
\text{Output} = AV,
\end{equation}
where $Q, K, V \in \mathbb{R}^{S\times D}$ and $M \in \{-\infty, 0\}^{S \times S}$ is a sparse mask managed implicitly by tile-based sliding. STA ensures $M$ only activates dense blocks, reducing redundant computation. This design ensures queries within a tile attend only to keys in predefined windows, eliminating irregular memory access patterns of traditional sliding window approaches. The number of dense blocks is:
\begin{equation}
S_{\text{dense}} = \left(\frac{W_T}{B_T} \times \frac{W_H}{B_H} \times \frac{W_W}{B_W} \right) \times \left(\frac{T}{B_T} \times \frac{H}{B_H} \times \frac{W}{B_W}\right).
\end{equation}
By aligning tile sizes with GPU thread blocks, STA minimizes masking overhead and maximizes hardware utilization, achieving up to $10.45\times$ speedup over full attention while maintaining generation quality. A key innovation of STA is its adaptability to head specialization, where different attention heads focus on varying spatial-temporal scales. Through profiling, STA automatically configures optimal window sizes per head, balancing computational efficiency with expressive power. However, STA relies on static window partitioning, which may not fully capture dynamic content-specific patterns, and requires careful tuning to avoid visual artifacts such as blocky boundaries.
\begin{figure*}[t]
\centering
\begin{minipage}{0.48\textwidth}
\begin{subfigure}{.47\linewidth}
\includegraphics[width=\linewidth]{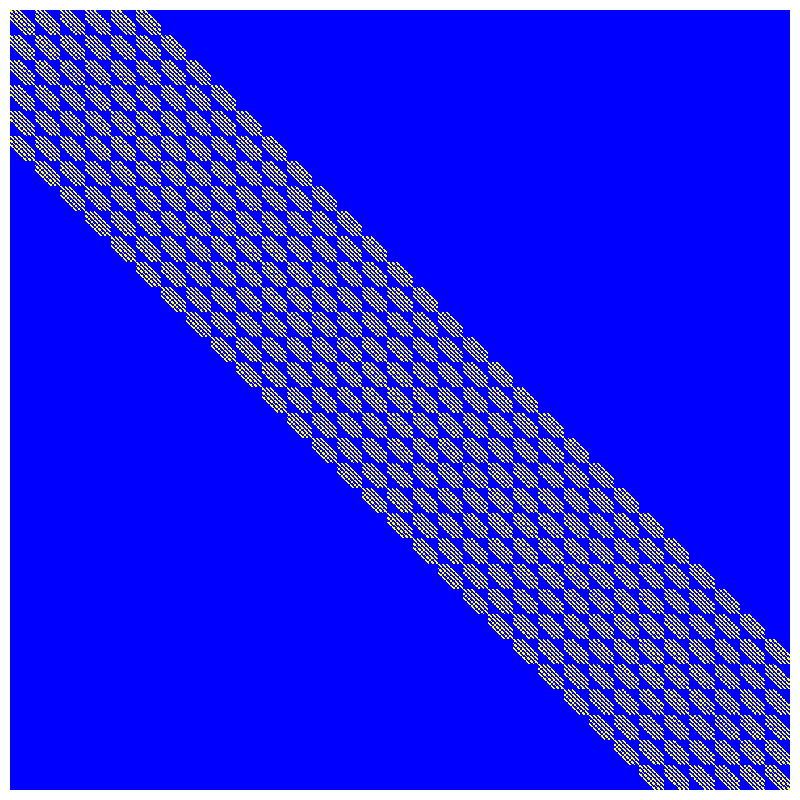}
\caption{STA (11, 40, 40)}
\label{fig:sta_11_40_40}
\end{subfigure}
\hfill
\begin{subfigure}{.47\linewidth}
\includegraphics[width=\linewidth]{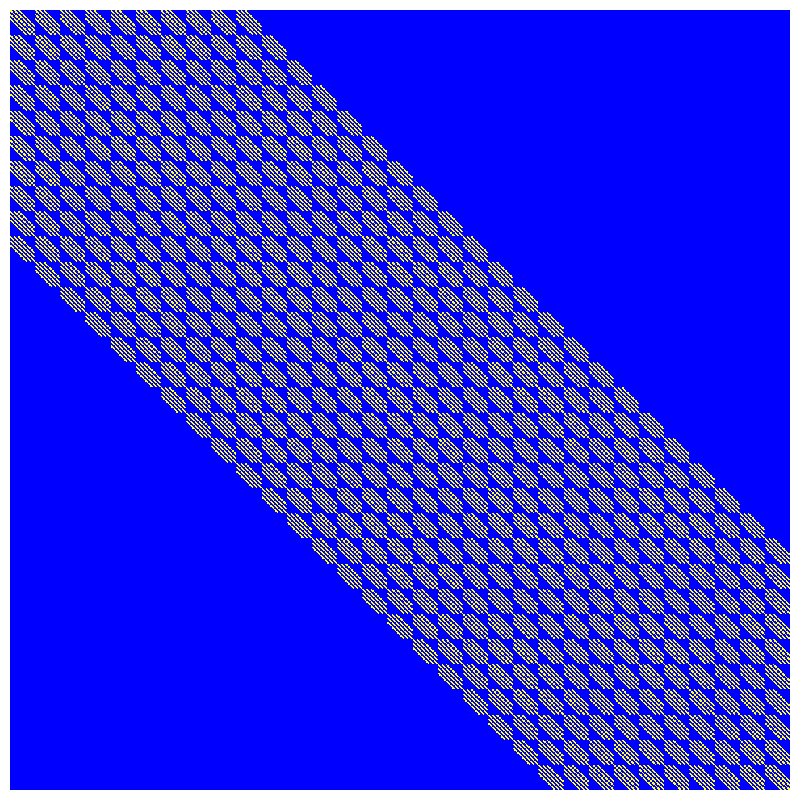}
\caption{STA (18, 40, 40)}
\label{fig:sta_18_40_40}
\end{subfigure}%

\begin{subfigure}{.48\linewidth}
\includegraphics[width=\linewidth]{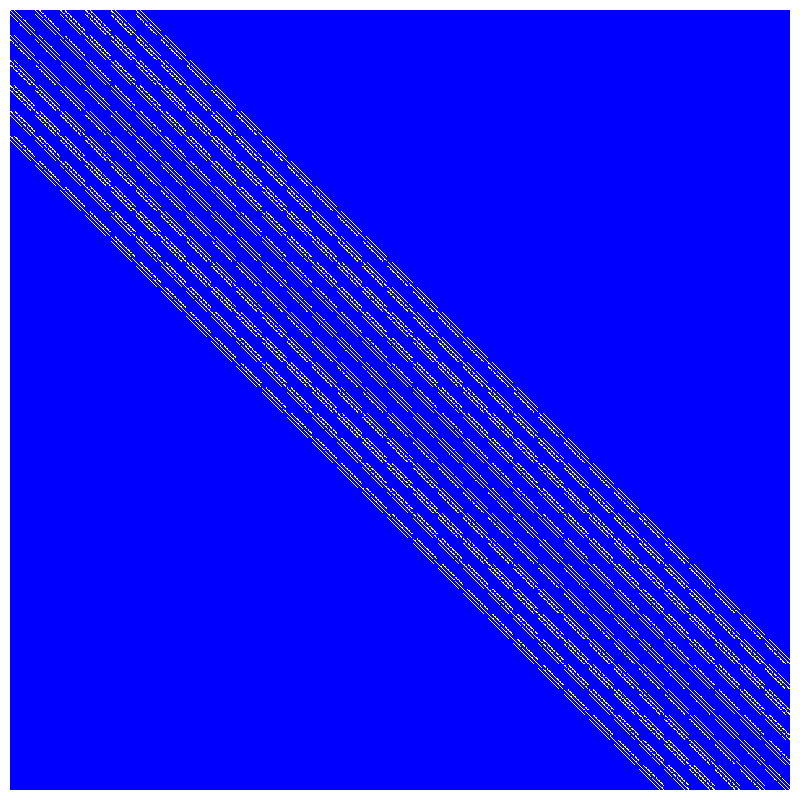}
\caption{STA (11, 24, 24)}
\label{fig:sta_11_24_24}
\end{subfigure}%
\hfill
\begin{subfigure}{.48\linewidth}
\includegraphics[width=\linewidth]{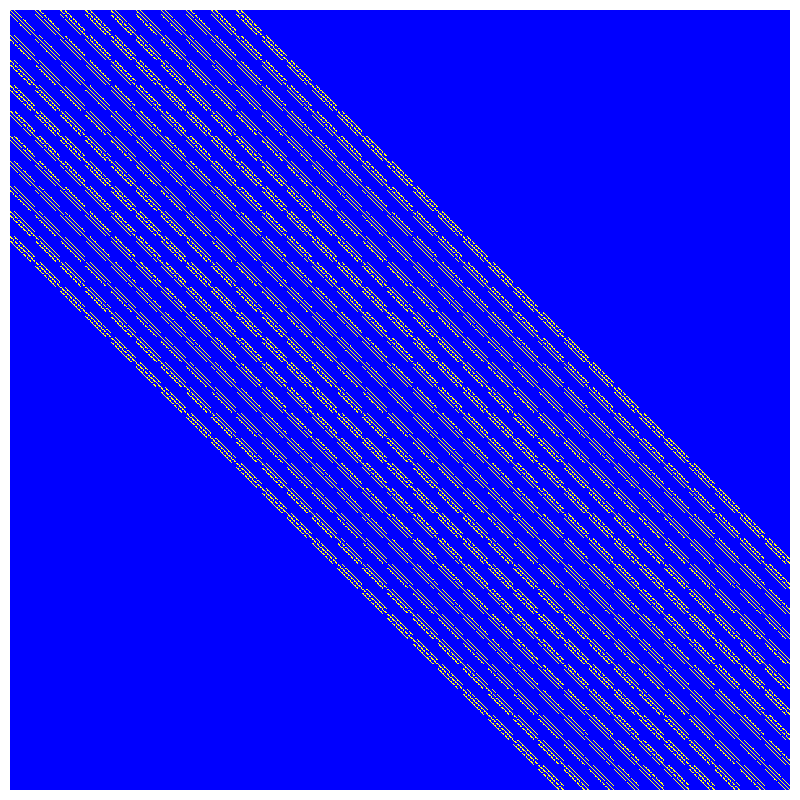}
\caption{STA (18, 24, 24)}
\label{fig:sta_18_24_24}
\end{subfigure}
\caption{STA masks with different window sizes.}
\label{fig:sta_masks}
\end{minipage}
\hfill
\begin{minipage}{0.4\textwidth}
\centering
\includegraphics[width=\linewidth]{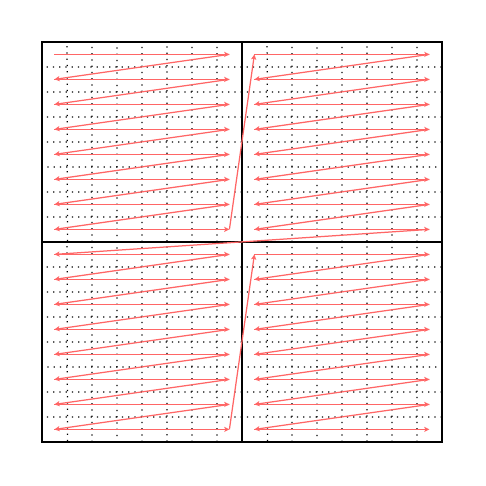}
\caption{Token reordering illustration for a latent image with height 16, width 16, and patch size 8. The diagram shows how spatial tokens are reorganized into fractal-flattened sequences while preserving their semantic relationships.}
\label{fig:token_reordering}
\end{minipage}
\end{figure*}
\subsection{Motivation for NABLA}
During our experiments with STA, we identified an essential issue of object duplication in high-resolution generation and long video sequences (see Examples 6 and 7 in Appendix~\ref{generation_examples}). This artifact appears in the case of non-optimal STA configuration and is caused by insufficient global attention coverage. Finding the optimal configuration is time-consuming, and as we show later, a universally optimal static configuration does not always exist. To address this, we hypothesized that an effective sparse attention algorithm must preserve long-range dependencies---connections between tokens distant in space or time. However, the semantics of such dependencies are inherently complex, making them infeasible to capture with fixed sparsity patterns. While STA provides significant efficiency gains through hardware-aware sparsity, its static nature limits adaptability to diverse video content. Figure~\ref{fig:attn_maps} contains examples (Fig.~\ref{fig:b}--\ref{fig:h}) of attention weights that differ significantly from the static STA pattern (Figure~\ref{fig:sta_masks}).

We argue that adaptive sparsity is essential: the algorithm must dynamically select sparse connections based on the actual input context. This motivates our proposed method, NABLA, which learns context-aware sparsity patterns to maintain global coherence. However, simply adopting dynamic sparsity introduces new challenges. On the one hand, purely adaptive methods can introduce visible border artifacts between areas recovered from neighboring latent pixels (see Examples 8 and 9 in Appendix~\ref{generation_examples}). On the other hand, existing dynamic frameworks often impose significant operational overheads that hinder their practical adoption for both training and inference.

Table~\ref{tab:method_comparison_motivation} summarizes the operational characteristics of state-of-the-art sparse attention strategies. Static methods like STA~\cite{zhang2025fast} incur negligible overhead but lack adaptability. Dynamic methods offer adaptability at a cost: MInference~\cite{jiang2024minference} requires an \textit{offline profiling} stage prior to inference to select the best fixed pattern from a dictionary, limiting its applicability to dynamic video contexts where content changes rapidly. Sparse VideoGen~\cite{xi2025sparse} employs \textit{online profiling} to classify heads into spatial or temporal types at every step, introducing non-trivial latency overhead within the generation loop and often requiring custom CUDA kernels. Other methods like AdaSpa~\cite{xia2025trainingfree} utilize hierarchical search, which can be computationally intensive.

\begin{table*}[t]
\centering
\caption{Operational comparison of sparse attention strategies. We analyze the source of overhead, implementation requirements, and adaptivity.}
\label{tab:method_comparison_motivation}
\resizebox{\textwidth}{!}{%
\begin{tabular}{lcccccc}
\toprule
\textbf{Method} & \textbf{Type} & \textbf{Profiling Overhead} & \textbf{Custom Kernels?}\textsuperscript{*} & \textbf{Mask Granularity} & \textbf{Training Support} & \textbf{FlexAttention?} \\
\midrule
Full Attention & Dense & No & No & Token-level & Yes & Yes \\
STA~\cite{zhang2025fast} & Static & Required (once) & No & Fixed Window & Yes & Yes \\
MInference~\cite{jiang2024minference} & Dynamic (Offline) & Required (Per-inference) & Yes & Pattern Dictionary & Limited & No \\
Sparse VideoGen~\cite{xi2025sparse} & Dynamic (Online) & Required (Per-step) & Yes & Head-Type (S/T) & Limited & No \\
AdaSpa~\cite{xia2025trainingfree} & Dynamic (Search) & High (Search) & No & Block-level & Limited & Partial \\
\textbf{NABLA (Ours)} & \textbf{Dynamic (Direct)} & \textbf{No} & \textbf{No} & \textbf{Block-level (Content-aware)} & \textbf{Yes} & \textbf{Yes} \\
\bottomrule
\end{tabular}%
}
\begin{minipage}{\textwidth}
\vspace{0.5em}
\small\textit{* While custom low-level kernels are an excellent means to maximize efficiency, their necessity significantly reduces model portability across diverse hardware platforms. To broaden accessibility and ease of integration, NABLA is intentionally designed to rely solely on standard PyTorch FlexAttention without mandatory custom kernels. However, this design choice does not preclude the future development of specialized kernels for NABLA.}
\end{minipage}
\end{table*}

In contrast, NABLA computes the mask \textit{online} via efficient block-wise pooling and CDF thresholding. This process introduces minimal arithmetic overhead (dominated by sorting small vectors) that scales linearly with the number of blocks and is fully integrated into the standard PyTorch execution graph without requiring custom CUDA kernels or external profiling stages. To address the boundary artifacts inherent to pure adaptive sparsity, we combine NABLA with STA, leveraging the strong prior of static windows while retaining the flexibility of dynamic selection. This hybrid approach achieves both computational efficiency and high-quality generation, addressing the limitations of purely static or dynamic approaches. By eliminating the need for specialized kernel development and profiling phases, NABLA is particularly suitable for dynamic video generation tasks where content varies significantly across frames and prompts, and where seamless integration into existing training pipelines is critical.
\section{Method}
\subsection{Training algorithm}
This section presents a training (or fine-tuning) version of NABLA algorithm. The method consists of the following parts:
\begin{itemize}
\item Token reordering to place tokens of the same spatial block to a continuous sequence
\item NABLA adaptive sparsification method that dynamically selects blocks of the feature map for which we perform attention computation
\item Sliding Tile Attention (STA) to improve fine-grained quality of the generated videos
\end{itemize}
\subsubsection{Token Reordering}
Following the approach of STA~\cite{zhang2025fast}, we find token reordering crucial for establishing semantic connections between adjacent tokens. Our method employs fractal flattening with spatial patches of size $P \times P$, which groups all tokens within each patch into a contiguous sequence of $P^2$ tokens. Notably, we preserve the original ordering along the temporal dimension.
Figure~\ref{fig:token_reordering} illustrates this transformation. We apply the reordering operation at the input stage of the DiT network and its inverse at the output stage, ensuring proper spatial relationships while maintaining computational efficiency.
This fractal flattening depends only on the patch size and is independent of global video dimensions. Since the pooling and thresholding operations in NABLA are performed locally within fixed blocks, the sparsity pattern adapts to local semantic structures rather than the total sequence length. This allows NABLA to generalize seamlessly across varying resolutions and aspect ratios without hyperparameter re-tuning. We empirically validated this scalability in our experiments (Section~\ref{sec:experiments}), demonstrating consistent efficiency and quality preservation. We employ 2D spatial flattening rather than a full 3D permutation because diffusion transformers are typically trained with mixed batches containing videos of varying lengths and standalone images. Our 2D approach treats the temporal dimension uniformly, making it applicable to both videos and single-frame images without requiring different processing logic for different data types.
\subsubsection{NABLA mask computation algorithm}
Algorithm~\ref{alg:nabla_t_v1_doc} presents the NABLA mask computation for Multi-Head Self-Attention. The algorithm takes as input a data sample represented by queries $Q$ and keys $K$, each containing $S$ tokens of dimension $D$ (we omit the batch dimension for simplicity). We denote the number of transformer heads as $h$ to distinguish it from the latent frame height $H$.
The core idea of our method involves computing a full attention map for downsampled versions of $Q$ and $K$, followed by binarization with minimal information loss. The downsampling is performed through average pooling of tokens in blocks of size $N=P^2$, making the reduced attention map computation $N^2$ times more efficient than computing the full attention map.
After computing the reduced attention map, we apply the softmax operation and compute the cumulative distribution function (CDF) for each row. We then binarize the map by retaining only values whose CDF exceeds the threshold $1 - thr$, where $thr$ is algorithm parameter. The binarization yields a unique sparsity pattern for each head, represented by an $S/N \times S/N$ matrix of binary values indicating whether to compute attention for the corresponding $N \times N$ block. Note that the $\mathrm{softmax}$, $\mathrm{sort}$, and $\mathrm{cumsum}$ operations are applied along the last dimension of the input tensor.
\begin{algorithm}[ht]
\caption{NABLA Sparse Mask Generation}
\label{alg:nabla_t_v1_doc}
\begin{algorithmic}[1]
\Statex \textbf{Require:}
\Statex \quad Query tensor: $Q \in \mathbb{R}^{h \times S \times D}$
\Statex \quad Key tensor: $K \in \mathbb{R}^{h \times S \times D}$
\Statex \quad Binarization threshold: $thr$
\Statex \quad Block size: $N$
\Statex \textbf{Reduced Attention Map Computation:}
\State $Q \gets \mathrm{reshape}(Q, [h, S/N, N, D])$
\State $K \gets \mathrm{reshape}(K, [h, S/N, N, D])$
\State $Q_a \gets \mathrm{mean}(Q, \mathrm{dim}=-2)$ \Comment{\parbox[t]{2.5cm}{Block averaging, $Q_a \in \mathbb{R}^{h \times S/N \times D}$}}
\State $K_a \gets \mathrm{mean}(K, \mathrm{dim}=-2)$ \Comment{\parbox[t]{2.5cm}{Block averaging, $K_a \in \mathbb{R}^{h \times S/N \times D}$}}
\State $K_a^T \gets K_a.\mathrm{transpose}(-2,-1)$ \Comment{$K_a^T \in \mathbb{R}^{h \times D \times S/N}$}
\State $A \gets \mathrm{softmax}(\frac{Q_a K_a^T}{\sqrt{D}})$ \Comment{\parbox[t]{3.5cm}{Reduced attention map, $A \in \mathbb{R}^{h \times S/N \times S/N}$}}
\Statex \textbf{Binarization via CDF:}
\State $vals, order \gets \mathrm{sort}(A)$ \Comment{Row-wise sorting}
\State $cvals \gets \mathrm{cumsum}(vals)$ \Comment{Cumulative sum}
\State $M \gets cvals \geq 1 - thr$ \Comment{Binarization of ordered values}
\State $M_{\nabla} \gets \mathrm{reorder}(M, order)$ \Comment{Original order restoration}
\State \Return $M_{\nabla}$
\end{algorithmic}
\end{algorithm}
\subsubsection{Joint NABLA and STA Sparsity Mask}
We find that combining NABLA with STA results in the best visual quality, benefiting from both our method's adaptive nature and STA's strong prior mask. The STA mask is computed as $M_{STA} = \mathrm{STA\_mask}(T, H, W, W_T, W_H, W_W)$, where:
\begin{itemize}
\item $T$: Number of latent frames in the video sample ($T=1$ for images)
\item $H,W$: Latent frame height and width
\item $W_T,W_H,W_W$: STA window parameters
\end{itemize}
The final mask for each data sample is given by $M = M_{\nabla} \lor M_{STA}$. After the mask is computed, it can be used directly in the Flex Attention~\cite{dong2024flexattentionprogrammingmodel} algorithm to improve the training efficiency.

\section{Experiments}
\label{sec:experiments}

\begin{table*}[t]
\centering
\begin{minipage}{0.48\linewidth}
\centering
\caption{Computational efficiency comparison. All measurements performed on 4$\times$ H100 GPUs.}
\label{tab:computation_efficiency}
\setlength{\tabcolsep}{1mm}
\begin{tabular}{lrrr}
\toprule
Method & Sparsity, & Inference time, \\
& \% & min \\
\midrule
Baseline & 0 & 8.35 \\
\midrule
STA(18,40,40) & 79.45 & 4.00 \\
NABLA(0.7) & 80.13 & 4.02 \\
NABLA(0.5)+STA(11,40,40) & 81 & 3.58 \\
\midrule
STA(18,24,24) & 91.28 & 3.08 \\
NABLA(0.4) & 92.5 & 3.07 \\
NABLA(0.2)+STA(11,24,24) & 92.27 & 3.13 \\
\bottomrule
\end{tabular}
\end{minipage}
\hfill
\begin{minipage}{0.48\linewidth}
\centering
\caption{Model quality metrics after fine-tuning. NABLA variants maintain comparable performance to baseline even at 90\% sparsity.}
\label{tab:finetuning_metrics}
\begin{tabular}{l|rrrrr}
        \toprule
        Method & CLIP & FVD &\multicolumn{3}{c}{VBench score$\uparrow$} \\
         & score$\uparrow$ & score$\downarrow$ & Quality  & Semantic  & Total  \\
        \midrule
        Wan2.1-14B\\ (Baseline) & 42.06 & 68.9 & \bf{85.15} & 75.23 & 83.16\\
        \midrule
        STA (18,24,24) & 41.51 & 74.06 & 85.05 & 71.73 & 82.39\\ \\
        NABLA(0.4) & \bf{42.08} & \bf{67.5} & 85.02 & 75.76 & 83.17\\ \\
        NABLA(0.2)+\\ STA(11,24,24) & 41.98 & 73.78 & 85.03 & \bf{76.04} & \bf{83.22}\\
        \bottomrule
\end{tabular}
\end{minipage}

\end{table*}\subsection{Fine-tuning experiments}

We evaluate our method in the fine-tuning setup using the Wan~2.1~14B T2V model~\cite{wan2025} at 720p resolution, focusing specifically on self-attention blocks due to their dominant contribution to overall FLOPs. We implement NABLA alongside STA as our baseline sparse attention method. For reproducibility, we use Flex Attention implementation from PyTorch 2.7. All experiments maintain consistent hardware and software configurations to ensure fair comparisons.

We perform knowledge distillation of the teacher model Wan2.1~T2V~14B using MSE loss. We initialize the student model with the baseline model weights and replace all self-attention blocks with sparse attention. Complete hyperparameter details are provided in Appendix~\ref{ft_hyperparameters}. Tables~\ref{tab:computation_efficiency} and~\ref{tab:finetuning_metrics} present our key findings.
Table~\ref{tab:computation_efficiency} presents the sparsity and acceleration settings for STA, NABLA, and STA+NABLA, which are used in subsequent quality evaluations.
Detailed evaluation results are provided in Table~\ref{tab:finetuning_results_detailed}.
\begin{table*}[t]
\centering
\begin{tabular}{@{}lcccc@{}}
\toprule
Metric & Wan2.1-14B  & STA & NABLA & NABLA(0.2)+\\
& & (18,24,24) & (0.4) &  STA(11,24,24) \\
\midrule
Subject consistency & 94.6 & 95.00 & \bf{95.01} & 93.18\\
Background consistency & 98.63 & 98.44 & \bf{98.67} & 98.25 \\
Aesthetic quality & 67.27 & 67.51 & 67.10 & \bf{67.63} \\
Imaging quality & \bf{66.46} & 66.18 & 66.35 & 66.28 \\
Object class & 81.09 & 82.12 & \bf{85.83} & 82.91 \\
Multiple objects & \bf{70.57} & 50.53 & 66.23 & 67.98 \\
Color & 89.83 & 85.61 & 85.19 & \bf{92.22} \\
Spatial relationship & 70.97 & 66.45 & \bf{75.44} & 70.07 \\
Scene & 45.36 & 48.11 & 50.07 & \bf{51.38} \\
Temporal style & 23.34 & 22.95 & \bf{23.46} & 22.99 \\
Overall consistency & 25.80 & \bf{26.65} & 26.10 & 26.00 \\
Human action & \bf{95} & 90 & 91 & 93 \\
Temporal flickering & \bf{98.91} & 98.83 & 98.88 & 98.78 \\
Motion smoothness & 98.38 & \bf{98.65} & 98.53 & 98.58 \\
Dynamic degree & 70.83 & 68.06 & 68.05 & \bf{72.22} \\
Appearance style & 22.69 & 22.51 & \bf{23.18} & 23.14 \\
Quality score & \bf{85.15} & 85.05 & 85.02 & 85.03 \\
Semantic score & 75.23 & 71.73 & 75.76 & \bf{76.04} \\
Total score & 83.16 & 82.39 & 83.17 & \bf{83.22} \\
\bottomrule
\end{tabular}
\caption{VBench results for 90\% sparsity. Bold values indicate the best performance in each category.}
\label{tab:finetuning_results_detailed}
\end{table*}
Key observations from our experiments include:
\begin{itemize}
\item NABLA achieves full quality recovery in generation metrics (CLIP, FVD and VBench scores)
\item The STA-only configuration shows degradation in VBench semantic scores
\item Detailed results (Table~\ref{tab:finetuning_results_detailed}) reveal STA's particular challenges with multiple objects and spatial relationships
\item Pure NABLA and NABLA+STA combinations maintain baseline-level performance across all objective metrics
\end{itemize}

\subsection{Human Evaluation}
We conducted a side-by-side human evaluation comparing generated video quality obtained by the baseline model and the finetuned models across various configurations. 50 participants evaluated 20 video pairs each comparing videos on three key perceptual dimensions (prompt alignment, visual quality, dynamics) selecting one option per dimension: left is better, right is better, both are good, both are bad. Results in Table~\ref{tab:human_evaluation} show NABLA's perceptual parity with baseline even at high sparsity.
\begin{table*}[t]
\centering
\caption{Human evaluation results (SBS test, estimated $N=500$). Values represent percentages with 95\% confidence intervals (CI). The overlapping CIs between Baseline and NABLA at 80\% sparsity indicate no statistically significant difference ($p > 0.05$).}
\label{tab:human_evaluation}
\begin{tabular}{l|cccc}
\toprule
\textbf{Winner} & \textbf{Semantic} & \textbf{Visual} & \textbf{Motion} & \textbf{Overall} \\
& \textbf{Alignment (\%)} & \textbf{Quality (\%)} & \textbf{Naturalness (\%)} & \textbf{(\%)} \\
\midrule
Baseline better 
& $19.1 \pm 3.5$ & $31.4 \pm 4.1$ & $10.5 \pm 2.7$ & $20.3 \pm 3.5$ \\
NABLA(0.7) better 
& $13.3 \pm 3.0$ & $26.7 \pm 3.9$ & $15.2 \pm 3.2$ & $18.4 \pm 3.4$ \\
Both good 
& $66.7 \pm 4.2$ & $40.0 \pm 4.3$ & $64.8 \pm 4.2$ & $57.2 \pm 4.3$ \\
Both bad 
& $\phantom{0}0.9 \pm 0.8$ & $\phantom{0}1.9 \pm 1.2$ & $\phantom{0}9.5 \pm 2.6$ & $\phantom{0}4.1 \pm 1.7$ \\
\bottomrule
\end{tabular}
\end{table*}

\textbf{Evaluation protocol:}
\begin{itemize}
\item \textbf{Video pairs:} Two 5-second preliminary generated videos for comparing methods are shown for randomly selected prompt (left/right video methods are shuffled).
\item \textbf{Prompts:} We use 942 diverse text prompts from VBench. The prompt for current videos is also visible to the user.
\item \textbf{Judge:} The user can watch the videos in repeat mode with the ability to zoom in/out and pause. The user must decide which video (left or right) is better, or both videos are good or bad.
\item \textbf{Dimensions:}
\begin{itemize}
\item Visual Quality: Artifact freedom and sharpness
\item Motion Naturalness: Better dynamics, physical plausibility and fluidity
\item Semantic Alignment: Prompt-video consistency
\end{itemize}
\end{itemize}
\subsection{Pretraining experiments}
\begin{figure}[ht]
\centering
\includegraphics[width=.8\linewidth]{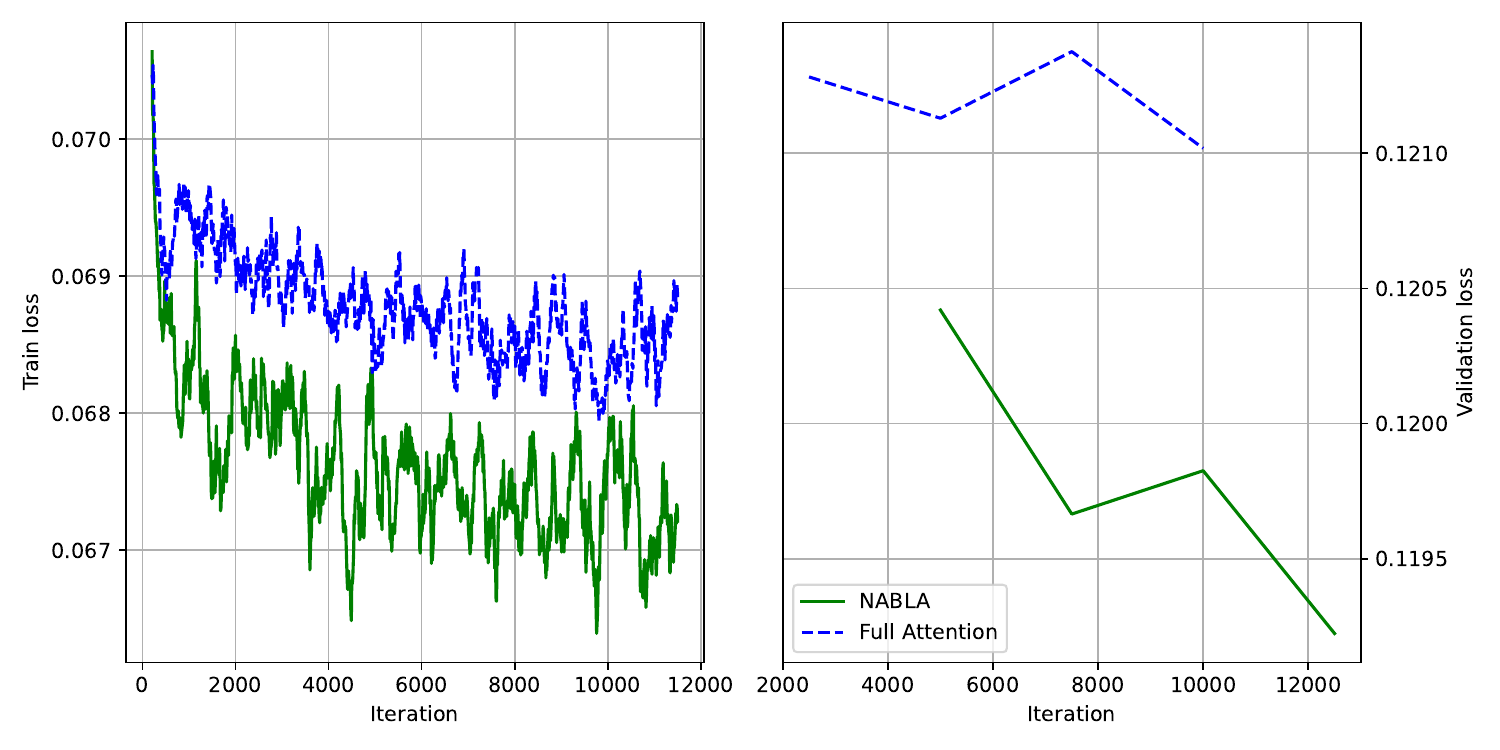}
\caption{Training convergence for full attention and NABLA models at 512$\times$512 resolution. NABLA achieves lower training and validation losses.}
\label{fig:pretrain_convergence}
\end{figure}

To verify the applicability of our method during pretraining, we train a custom DiT-based 2B model in three stages:
\begin{enumerate}
\item Text-to-image pretraining at 256$\times$256 resolution with full attention.
\item Text-to-video pretraining at 256$\times$256 resolution with full attention.
\item Text-to-video pretraining at 512$\times$512 resolution:
\begin{enumerate}
\item With full attention.
\item With the NABLA method (80\% sparsity).
\end{enumerate}
\end{enumerate}
The first stage is conducted from scratch, with each subsequent stage initialized using weights from the previous stage. The first two stages are common to all experiments.
We compare the convergence of training and validation losses between stages 3(a) and 3(b). Figure~\ref{fig:pretrain_convergence} shows that the NABLA model achieves better convergence than its full attention counterpart. Furthermore, each training iteration takes 10.9 seconds for the full attention model compared to 7.5 seconds for NABLA, resulting in a 1.46$\times$ speedup.
\section{Conclusion}
By dynamically adapting to sparsity patterns through block-wise attention with adaptive thresholding, our method achieves:
\begin{itemize}
    \item \textbf{Significant computational efficiency:} Empirically demonstrated speed-ups of $2.7\times$ for inference and $1.46\times$ for pre-training on specific high-resolution configurations (Wan 2.1 14B and custom 2B DiT), with potential for further optimization depending on sparsity settings.
    \item \textbf{Minimal quality degradation:} Demonstrates near-identical performance to full attention across quantitative metrics (CLIP, FVD, VBench) and human evaluations, confirming that dynamic sparsity preserves semantic fidelity.
    \item \textbf{Hardware-agnostic implementation:} Ensures seamless integration with PyTorch's FlexAttention without requiring custom CUDA kernels or auxiliary losses, facilitating immediate adoption in existing pipelines.
\end{itemize}Extensive experiments demonstrate NABLA's superiority over static sparsity approaches like STA, particularly in preserving long-range dependencies and handling complex spatial-temporal relationships. Our hybrid approach combining NABLA with STA further enhances visual quality by mitigating boundary artifacts while maintaining efficiency.
The proposed approach establishes a new state-of-the-art for efficient video generation, enabling high-resolution synthesis with reduced computational demands.

\bibliographystyle{ieeetr}
\bibliography{references}

\appendices
\section{Fine-Tuning Hyperparameters} \label{ft_hyperparameters}
We performed knowledge distillation on the full-attention Wan2.1 open-source model using MSE loss. The training utilized a specially curated dataset of high-quality, high-dynamic videos and high-quality images. The experiments were conducted on 256 H100 GPUs, using the following hyperparameters:
\begin{itemize}
\item Total batch size: 64
\item Sequence parallel: 4
\item Training steps: 1600
\item Optimizer: AdamW with:
\begin{itemize}
\item Learning rate: 1e-6
\item Weight decay: None
\item Betas: (0.9, 0.95)
\item Epsilon: 1e-8
\end{itemize}
\item Gradient norm: 0.01
\end{itemize}
\section{Generation Examples} \label{generation_examples}
\begin{figure}[!ht]
\centering
\includegraphics[width=.9\linewidth]{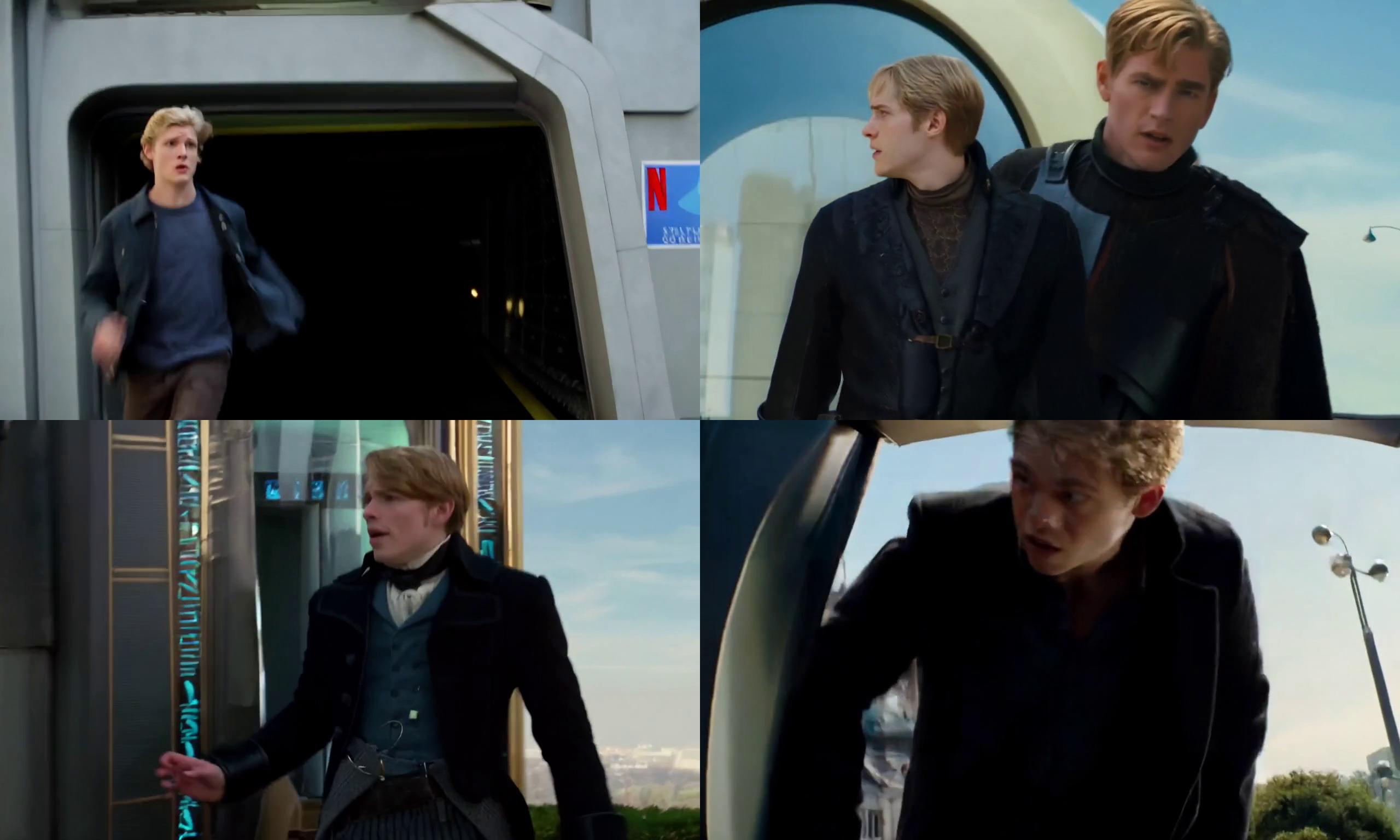}
\caption{Input prompt: \textit{Movie scene of a time portal opening up in a modern city, a 18th century young blonde man walks out of it looking confused, close-up, sci-fi, Netflix Original, professionally color graded, 35mm film}
}
\label{fig:examle_doubles1}
\end{figure}

\begin{figure}[!ht]
\centering
\includegraphics[width=.9\linewidth]{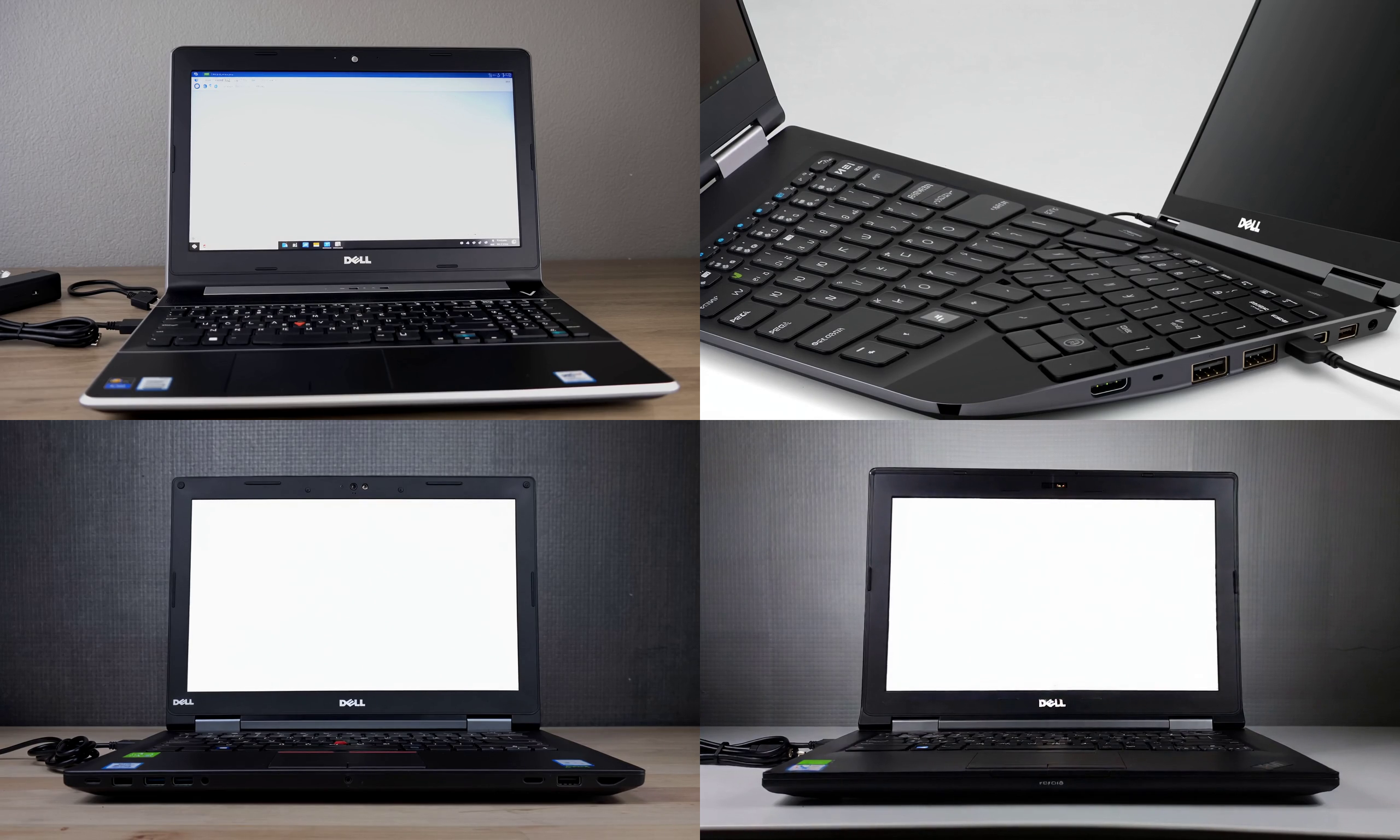}
\caption{Input prompt: \textit{A sleek black laptop made of durable aluminium with a flat rectangular shape. It is a medium-sized device with a 14-inch screen. The laptop features a backlit keyboard and comes with a charger. The text on the device reads 'Dell.'}
}
\label{fig:examle_doubles2}
\end{figure}

\begin{figure}[!ht]
\centering
\includegraphics[width=1.0\linewidth]{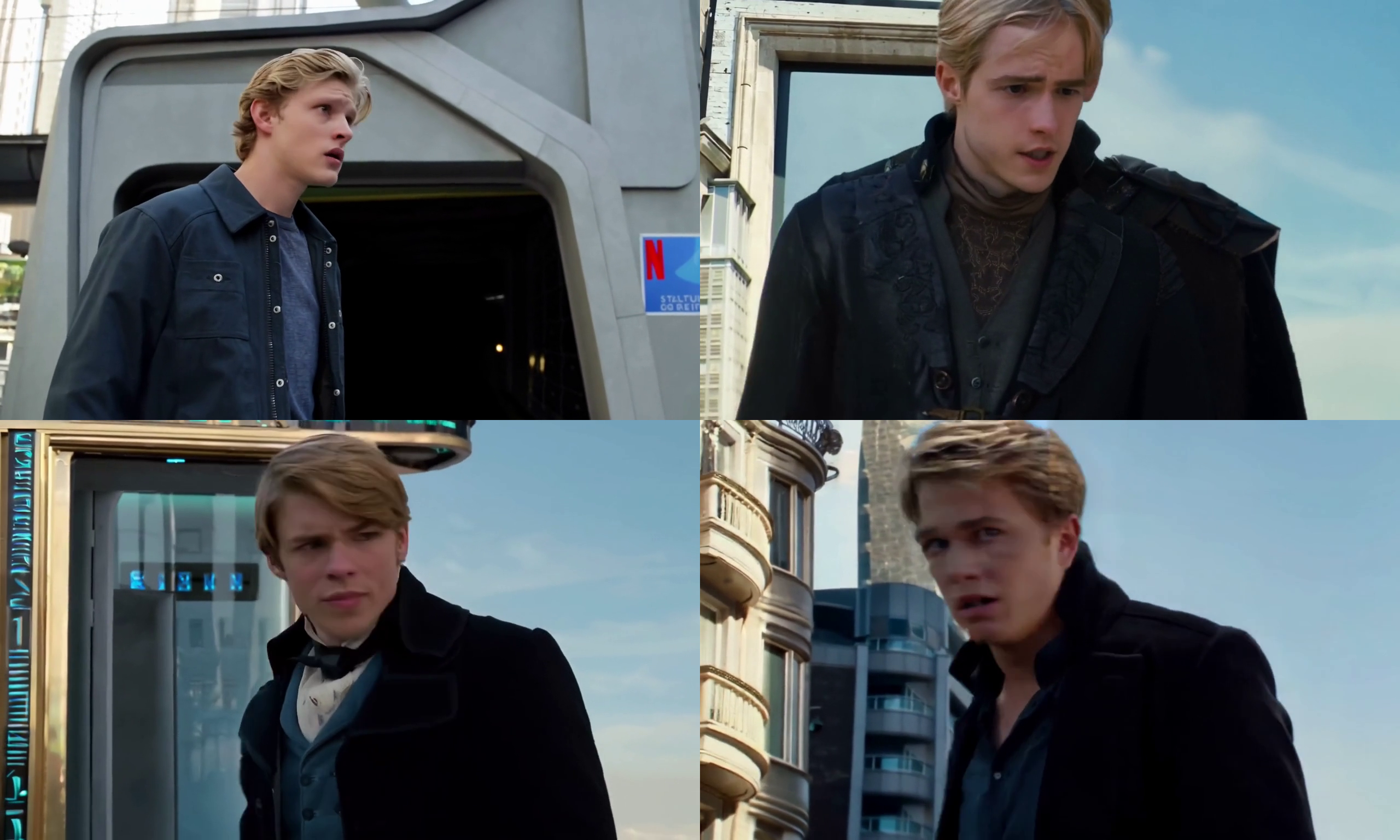}
\caption{Input prompt: \textit{Movie scene of a time portal opening up in a modern city, a 18th century young blonde man walks out of it looking confused, close-up, sci-fi, Netflix Original, professionally color graded, 35mm film}
}
\label{fig:examle_edges1}
\end{figure}

\begin{figure}[!ht]
\centering
\includegraphics[width=1.0\linewidth]{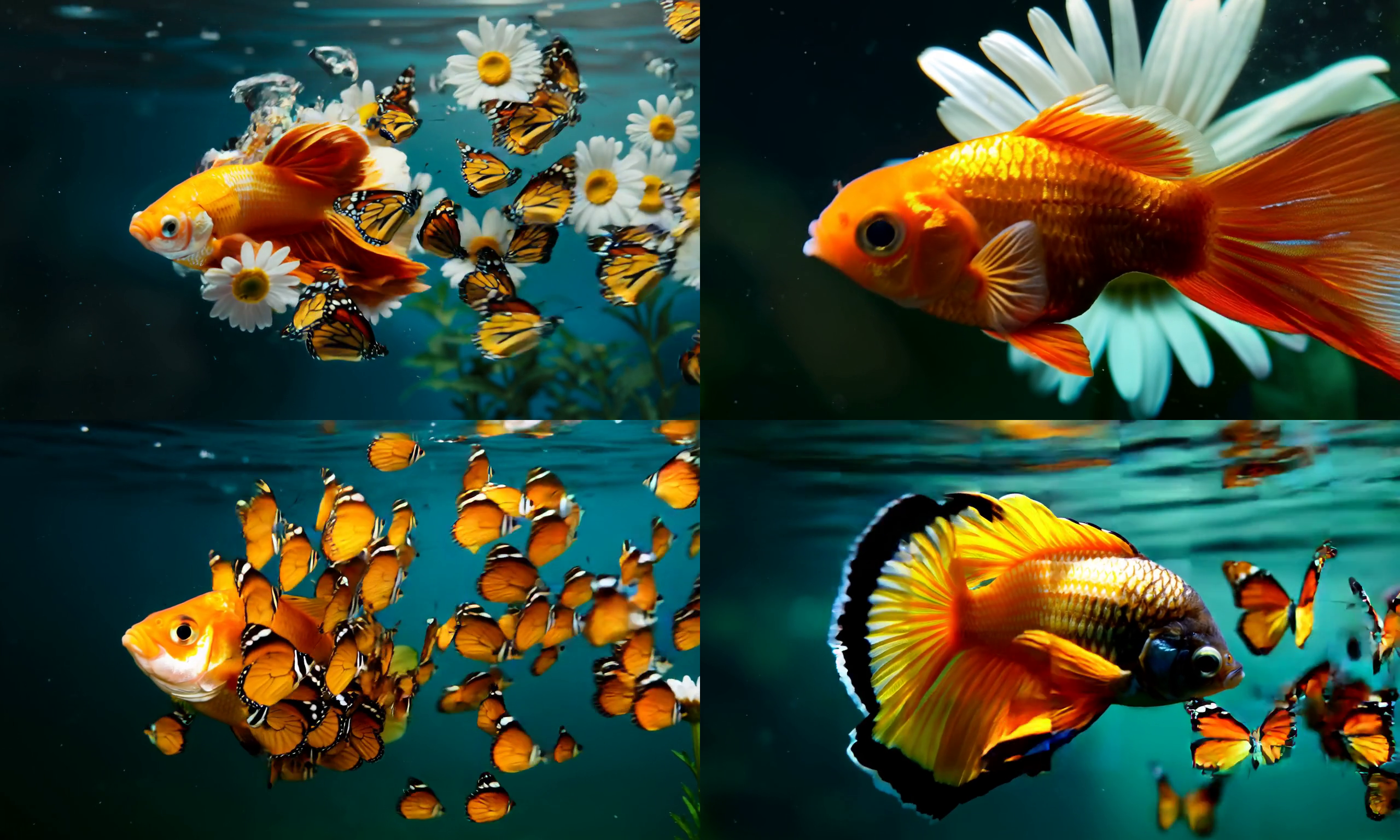}
\caption{Input prompt: \textit{"Cinematic shot of a Beta fish swimming, moving dynamically in the water. A daisy is transformed into a group of butterflies. The fish has orange, blue and yellow colors. Beautiful nature documentary, low contrast, 35mm, color correction.}
}
\label{fig:examle_edges2}
\end{figure}

The following examples (Fig.~\ref{fig:examle_doubles1},\ref{fig:examle_doubles2},\ref{fig:examle_edges1},\ref{fig:examle_edges2}) demonstrate generation results from different model configurations, presented clockwise from the top-left corner: Full Attention (Baseline Wan 2.1), STA(18,24,24), NABLA(0.4), and NABLA(0.2) + STA(11,24,24).
\section{Appendix: Theoretical Justification for CDF-Based Sparsification}
\label{app:cdf_theory}
\subsection{Motivation and Empirical Observations}
In video diffusion transformers (DiTs), the self-attention mechanism operates over a spatiotemporal token grid of size $T \times H \times W$, yielding a total sequence length $S = T \cdot H \cdot W$, making full attention computationally prohibitive.
Empirical analysis of attention maps (see Figure 2 in the main text) reveals three consistent properties:
\begin{enumerate}
\item \textbf{Locality}: Query tokens primarily attend to spatially and temporally nearby keys.
\item \textbf{Long-range anchors}: Certain heads exhibit strong attention to a fixed reference frame (e.g., the first frame of video/scene or another key frame), preserving object identity across time.
\item \textbf{Heterogeneous structure}: Attention patterns vary significantly across heads, timesteps, and video content.
\end{enumerate}
These properties induce a characteristic \textit{rapid decay} in sorted attention weights: a small set of keys captures most of the probability mass, while the tail decays near-exponentially. This behavior is particularly pronounced in long videos, where object motion or scene dynamics create diverse, non-stationary attention profiles. Consequently, a fixed sparsity budget (e.g., Top-$k$) is suboptimal: it either over-computes for sharply peaked distributions or under-approximates diffuse ones.
\subsection{Model Assumptions and Notation}
Let each row $\mathbf{a}^{(i)}$ of the attention matrix $A \in \mathbb{R}^{S \times S}$ belong to the probability simplex $\Delta^{S-1} = \{ \mathbf{p} \in \mathbb{R}^S : p_j \geq 0,\; \sum_j p_j = 1\}$. Following the empirical observations in Figure~2, we model the sorted attention weights for query $i$ as a geometric decay:
\begin{equation}
a^{(i)}_{\pi_i(j)} = (1 - \rho_i) \rho_i^{j-1}, \quad j = 1, \dots, S,
\end{equation}
where $\pi_i$ is the permutation sorting $\mathbf{a}^{(i)}$ in descending order, and $\rho_i \in [\rho_{\min}, \rho_{\max}] \subset (0, 1)$ is a latent decay-rate variable. Assuming $S \gg 1$ and $\rho_i < 1$, the truncation error $\rho_i^S$ is negligible, so rows are effectively normalized. Smaller $\rho_i$ implies stronger locality and higher potential sparsity. The parameter $\rho_i$ varies across queries and heads, reflecting the heterogeneity observed in practice (Figure~2).

We consider two sparsification strategies:
\begin{itemize}
    \item \textbf{Top-$k$}: Retain the top-$k$ keys. The $\ell_1$ error is $\varepsilon_i^{\text{top}} \approx \rho_i^k$.
    \item \textbf{CDF($\tau$)}: Retain the smallest prefix covering mass $1-\tau$, i.e., 
    \[
    r_i = \min\Bigl\{j : \sum_{t=1}^j a^{(i)}_{\pi_i(t)} \geq 1 - \tau\Bigr\}.
    \]
    The error satisfies $\varepsilon_i^{\text{cdf}} \leq \tau$.
\end{itemize}
We impose a \textit{fixed average sparsity budget}: $\mathbb{E}[r_i] = k$, where the expectation is taken over the distribution of $\rho_i$, assumed non-degenerate ($\mathrm{Var}(\rho_i) > 0$).

\subsection{Theorem and Proof}

\begin{theorem*}
Assume the attention matrix $A$ satisfies the geometric decay model described (with $\rho_i \in [\rho_{\min}, \rho_{\max}] \subset (0,1)$ and $\mathrm{Var}(\rho_i) > 0$). Then, for any fixed integer $k \geq 1$, there exists a unique threshold $\tau \in (0,1)$ such that $\mathbb{E}[r_i] = k$, and the expected $\ell_1$ approximation error of CDF sparsification is strictly smaller than that of Top-$k$:
\[
\mathbb{E}[\varepsilon^{\text{cdf}}] < \mathbb{E}[\varepsilon^{\text{top-}k}].
\]
\end{theorem*}

\begin{proof}
Under the assumptions, the tail errors are
\[
\varepsilon_i^{\text{top}} = \sum_{j=k+1}^S a^{(i)}_{\pi_i(j)} = \rho_i^k - \rho_i^S \approx \rho_i^k,
\qquad
\varepsilon_i^{\text{cdf}} = \rho_i^{r_i},
\]
where $r_i = \min\{j : 1 - \rho_i^j \geq 1 - \tau\} = \lceil \log_{\rho_i} \tau \rceil$, hence $\varepsilon_i^{\text{cdf}} \leq \tau$.

The function $\tau \mapsto \mathbb{E}[r_i(\tau)]$ is strictly decreasing and continuous from $(0,1)$ onto $(1,\infty)$, so there exists a unique $\tau$ satisfying $\mathbb{E}[r_i] = k$.

Now suppose, for contradiction, that $\mathbb{E}[\varepsilon^{\text{top}}] \leq \tau$. Then $\rho_i^k \leq \tau$ almost surely; otherwise the expectation would exceed $\tau$. This implies $\rho_i \leq \tau^{1/k}$ almost surely. Since $\mathrm{Var}(\rho_i) > 0$, the inequality is strict on a set of positive measure, so $r_i = \lceil \log_{\rho_i} \tau \rceil < k$ on that set. Because $r_i \leq k$ everywhere and $r_i < k$ with positive probability, it follows that $\mathbb{E}[r_i] < k$, contradicting the sparsity budget $\mathbb{E}[r_i] = k$.

Therefore, $\mathbb{E}[\varepsilon^{\text{top}}] = \mathbb{E}[\rho_i^k] > \tau \geq \mathbb{E}[\varepsilon^{\text{cdf}}]$, completing the proof.
\end{proof}
\subsection{Discussion and Practical Implications}
This result provides a theoretical foundation for NABLA's use of CDF-based sparsification: under realistic attention decay profiles, it achieves lower average error than fixed Top-$k$ at the same computational budget.
However, several caveats apply:
\begin{itemize}
\item During end-to-end training, the attention distribution may adapt to the sparsification mechanism itself (e.g., becoming more uniform under Top-$k$), potentially reducing the gap.
\item The exponential decay model is an idealization; real attention may exhibit multimodality or long tails.
\item The practical benefit depends on implementation overhead (e.g., CDF requires sorting or sampling), though NABLA mitigates this via block-level aggregation.
\end{itemize}
Thus, while the theory motivates CDF as a principled choice, its superiority must be validated empirically, as demonstrated in Tables 1 - 4 and Figure 2 of the main text.

\begin{IEEEbiography}[{\includegraphics[width=1in,height=1.25in,clip,keepaspectratio]{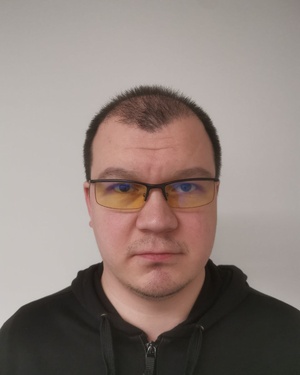}}]{Dmitrii Mikhailov}
Dmitrii Mikhailov graduated from the Department of Mathematical Theory of Intelligent Systems at Lomonosov Moscow State University in 2011. He subsequently worked as a researcher and developer at Neurocom and Huawei companies, where he focused on the practical application of machine learning to a broad range of tasks. Since 2019, his work has centered on hardware-aware algorithm optimization. Since 2024, he has been with Kandinsky Lab, where he specializes in optimizing the Kandinsky family of generative models.
\end{IEEEbiography}

\begin{IEEEbiography}[{\includegraphics[width=1in,height=1.25in,clip,keepaspectratio]{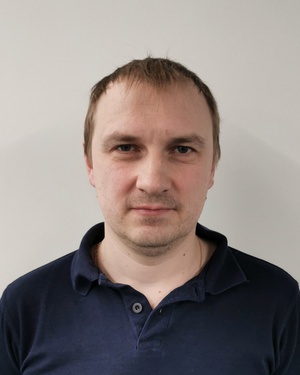}}]{Aleksey Letunovskiy} received M.S. and PhD degree from in Department of Mathematical Theory of Intelligent Systems at Lomonosov Moscow State University in 2005 and 2015. He subsequently worked as a researcher and developer at LSI Logic and Huawei companies, where he focused on the practical application of descrete mathematics and machine learning to a broad range of tasks. Since 2024, he has been with Kandinsky Lab, where he specializes in optimizing the Kandinsky family of generative models.
\end{IEEEbiography}

\begin{IEEEbiography}[{\includegraphics[width=1in,height=1.25in,clip,keepaspectratio]{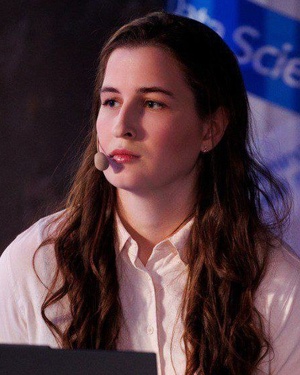}}]{Maria Kovaleva}
  received B.S. and M.S. degrees (Hons.) in applied mathematics and physics from the Moscow Institute of Physics and Technology (MIPT), Russia, in 2022 and 2024, respectively, and an M.S. degree (Hons.) in data science from the Skolkovo Institute of Science and Technology (Skoltech), Russia, in 2024.

From 2021 to 2022, she was an Engineer at the Laboratory of Wave Processes and Control Systems, MIPT, where she worked on a project focused on forecasting turning points in financial markets using reinforcement learning algorithms. In 2023, she was an Intern at the Samsung AI Research Center (SAIC Moscow), working on a speech enhancement project. From 2023 to 2024, she served as a Junior Research Engineer at the Applied Research Laboratory «Skoltech-Sberbank», contributing to several research projects dedicated to enhancing embeddings of time series data. Since 2024, she has been a Research Engineer at Kandinsky Lab, working in the field of generative computer vision and video generation, with a specific focus on the development of Kandinsky models.
\end{IEEEbiography}

\begin{IEEEbiography}[{\includegraphics[width=1in,height=1.25in,clip,keepaspectratio]{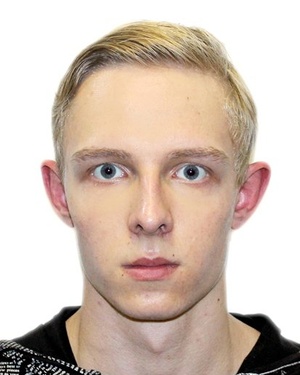}}]{Vladimir Arkhipkin}
received his B.S. and M.S. degree in Applied Mathematics and Physics from Moscow Institute of Physics and Technology (MIPT), Russia, in 2020 and 2022, respectively. From 2018 to 2020, Arkhipkin worked as a research assistant at the Landau Institute for Theoretical Physics, Russia. From 2022 to 2025, he was a data science researcher at Sber AI Lab, Sberbank Russia. Since 2025, he is a research direction leader, key researcher, and lead architecture developer at Kandinsky Lab, working on image and video generation with a focus on accelerating diffusion models, efficient attention mechanisms, and maintaining high generation quality during model pre-training and fine-tuning.
\end{IEEEbiography}

\begin{IEEEbiography}[{\includegraphics[width=1in,height=1.25in,clip,keepaspectratio]{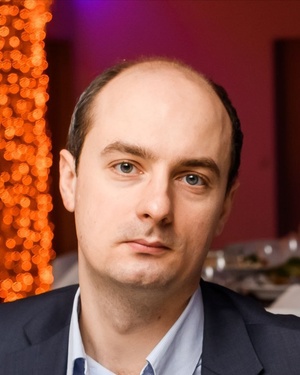}}]{Vladimir Korviakov} graduated with honors from Bauman Moscow State Technical University with a Specialist degree (Master's equivalent) in Flight Vehicle Control Systems. He later worked at the S.P. Korolev Rocket and Space Corporation Energia, where he contributed to the development of human-machine interfaces for piloted spacecraft and crew training simulators.

From 2018 to 2024, he was a researcher at Huawei's Moscow Research Center, focusing on computer vision, deep learning, multimodal data processing, and generative artificial intelligence models. Since 2019, he has led a team of research engineers specializing in the acceleration and optimization of neural networks for both training and inference, including models such as ResNet, Vision Transformer, CLIP, LLaMA, and Stable Diffusion.

He currently leads a team at Kandinsky Lab focused on optimizing diffusion models for efficient video and image generation. He is the inventor of six patents in the fields of neural networks and digital image processing.
\end{IEEEbiography}

\begin{IEEEbiography}[{\includegraphics[width=1in,height=1.25in,clip,keepaspectratio]{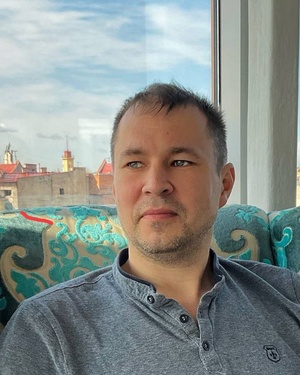}}]{Vladimir Polovnikov}
has a PhD in Mathematics, is a Senior Researcher at the Faculty of Mechanics and Mathematics, Lomonosov Moscow State University. His dissertation (2007) focused on the functional theory of neural networks, addressing architectural equivalence and computational complexity.
His research interests span structural synthesis of neural architectures, graph theory, energy-efficient computing for microprocessor systems, and coding theory. During his tenure at Huawei, he contributed to LDPC code development for 5G standards. At Kandinsky Lab, he develops diffusion models for image and video generation. His research efforts are also directed toward computational chemistry and biology, with a focus on molecular modeling and related neural network applications. For over 15 years, he has taught courses on neural networks, theory of discrete functions, and VLSI synthesis, while supervising interdisciplinary projects at the intersection of AI, materials science, and biomedicine.
\end{IEEEbiography}

\begin{IEEEbiography}[{\includegraphics[width=1in,height=1.25in,clip,keepaspectratio]{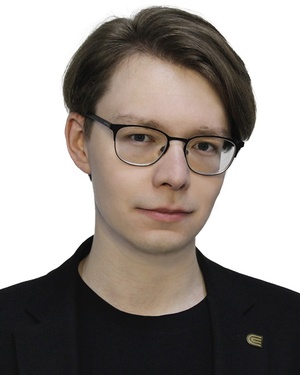}}]{Viacheslav Vasilev}
received his B.S. and M.S. degree (Hons.) in Applied Mathematics and Physics from Moscow Institute of Physics and Technology (MIPT), Russia, in 2021 and 2023, respectively. He also received M.S. degree (Hons.) in Data Science at the Skolkovo Institute of Science and Technology (Skoltech), Russia, in 2023.

In 2020, he was a Research Intern in deep learning and computer vision at the European Bioinformatics Institute (EMBL-EBI), Hinxton, UK, and in 2021 at Helmholtz-Zentrum Dresden-Rossendorf (HZDR), Dresden, Germany. From 2020 to 2022, he was a Junior Researcher at the Vision Systems Laboratory, Institute for Information Transmission Problems of the Russian Academy of Sciences. There, he worked on developing algorithms to improve the color reproduction quality of mobile phone cameras. In 2022, he worked on 3D generative computer vision at the Samsung AI Center. Since 2023, he has been a Researcher in generative computer vision and video generation at Kandinsky Lab, specifically focusing on the development of Kandinsky models.

Vasilev has experience lecturing at several international machine learning summer schools, including SMILES 2025 (China, Harbin). He is an author of a number of scientific publications, including top journals (Q1/Q2) and proceedings of Core A/A* conferences.
\end{IEEEbiography}

\begin{IEEEbiography}[{\includegraphics[width=1in,height=1.25in,clip,keepaspectratio]{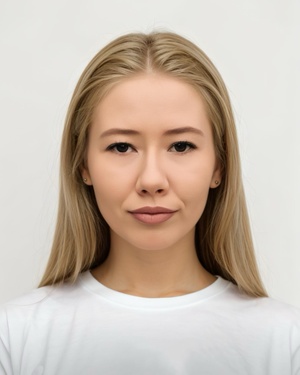}}]{Evelina Sidorova}
received her Bachelor's and Master's degrees in Information Systems and Technologies from Bauman Moscow State Technical University (BMSTU).

In her Master's thesis, she developed a system for generating images from textual descriptions based on a diffusion generative model. The work focused on the research and implementation of an architecture capable of synthesizing high-quality images that match the semantics of the input prompt.

Her research interests include: generative diffusion models, prompt engineering techniques with a focus on applications in business processes, as well as the study of world models within modern approaches to artificial intelligence and autonomous decision-making.
\end{IEEEbiography}

\begin{IEEEbiography}[{\includegraphics[width=1in,height=1.25in,clip,keepaspectratio]{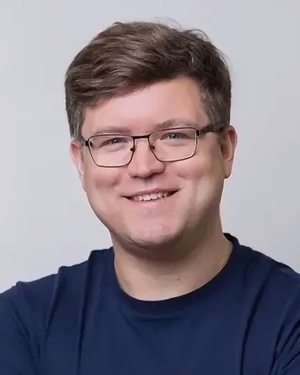}}]{Denis Dimitrov}
received his Specialist degree (equivalent to M.S. degree) in Mathematics and Mechanics from the Department of Probability Theory, Faculty of Mechanics and Mathematics, Lomonosov Moscow State University, Russia, in 2018. He completed his PhD studies at the Lomonosov Moscow State University from 2018 to 2022, focusing on strictly mathematical issues concerning statistical estimation of the f-divergences and its applications, such as feature selection and multivariate in homogeneities detection. Since 2020 and until 2025, Dimitrov led the Sber AI Research laboratory and served as a scientific advisor at the Artificial Intelligence Research Institute (AIRI). Under his leadership, AIRI conducted the first experiments in Russia on creating multimodal models, which ultimately led to the development of OmniFusion, the country's first dialogue visual language model. His current research focuses on generative modelling: text-to-image and text-to-video models, large language and multimodal models, and related directions.
Currently, Dimitrov heads Kandinsky Lab, where he continues to advance research in efficient generative AI. He is the founder and leader of the Kandinsky line of image and video generation models, and one of the co-creators of the GigaChat series of large language models. Dimitrov is an Associate Professor and lecturer at Lomonosov Moscow State University and regularly presents at international scientific and technical conferences. He is the author of numerous scientific publications, including papers in top-tier journals (Q1/Q2) and proceedings of Core A/A* conferences, h-index=16.
\end{IEEEbiography}
\EOD
\end{document}